\documentclass{article} 
\usepackage{iclr2019_conference,times}

\usepackage{amsmath,amsfonts,bm}

\def\eqref#1{equation~\ref{#1}}

\def\1{\bm{1}}

\def\eps{{\epsilon}}

\DeclareMathAlphabet{\mathsfit}{\encodingdefault}{\sfdefault}{m}{sl}
\SetMathAlphabet{\mathsfit}{bold}{\encodingdefault}{\sfdefault}{bx}{n}

\DeclareMathOperator*{\argmax}{arg\,max}
\DeclareMathOperator*{\argmin}{arg\,min}

\usepackage[english]{babel}
\usepackage[utf8]{inputenc} 
\usepackage[T1]{fontenc}    
\usepackage{url}            
\usepackage{booktabs}       
\usepackage{amsfonts}       
\usepackage{nicefrac}       
\usepackage{microtype}      
\usepackage{graphicx}
\usepackage{amsmath}
\usepackage{amssymb}
\usepackage{subfigure}
\usepackage{epsfig}
\usepackage{verbatim}
\usepackage{multirow}
\usepackage{color}
\usepackage{algorithm}
\usepackage{algpseudocode}
\usepackage{mathptmx}
\usepackage{txfonts}
\usepackage{wrapfig}
\usepackage{bbm}

\title{Towards Metamerism via Foveated Style Transfer}

\author{
  Arturo Deza$^{1,4}$, Aditya Jonnalagadda$^{3}$, Miguel P. Eckstein$^{1,2,4}$ \\
  $^1$ Dynamical Neuroscience, $^2$Psychological and Brain Sciences,
  \\ $^3$Electric and Computer Engineering, $^4$ Institute for Collaborative Biotechnologies\\
  UC Santa Barbara, CA, USA\\
  \tt\small{deza@dyns.ucsb.edu,aditya\_jonnalagada@ece.ucsb.edu,eckstein@psych.ucsb.edu} \\
}

\begin{document}
\iclrfinalcopy 

\maketitle

\begin{abstract}
The problem of \textit{visual metamerism} 
is defined as finding a family of perceptually indistinguishable, yet physically 
different images.
In this paper, we propose our NeuroFovea metamer model, a foveated generative model
that is based on a mixture of peripheral representations and 
style transfer forward-pass algorithms. 
Our gradient-descent free model is parametrized by a foveated VGG19 encoder-decoder which allows us 
to encode images in high dimensional space and interpolate between the content and texture information with adaptive instance normalization anywhere in the visual field.
Our contributions include: 1) 
A framework for computing metamers that resembles a noisy communication 
system via a foveated feed-forward encoder-decoder network -- 
We observe that metamerism arises as a byproduct of noisy 
perturbations that partially lie in the perceptual null space;
2) A perceptual optimization scheme as a solution to the hyperparametric nature of our metamer model that
requires tuning of the image-texture tradeoff coefficients everywhere in the visual field which are a consequence of internal noise; 
3) An ABX psychophysical evaluation of our metamers where we also find that the rate of growth
of the receptive fields in our model match V1 for reference metamers and 
V2 between synthesized samples.
Our model also renders metamers at roughly a second, presenting a $\times1000$ speed-up compared to the previous work,
which allows for tractable data-driven metamer experiments.
\end{abstract}

\vspace{-10pt}
\section{Introduction}
\begin{figure}[t]
\centering
\includegraphics[width=1.0\columnwidth,clip=true,draft=false,]{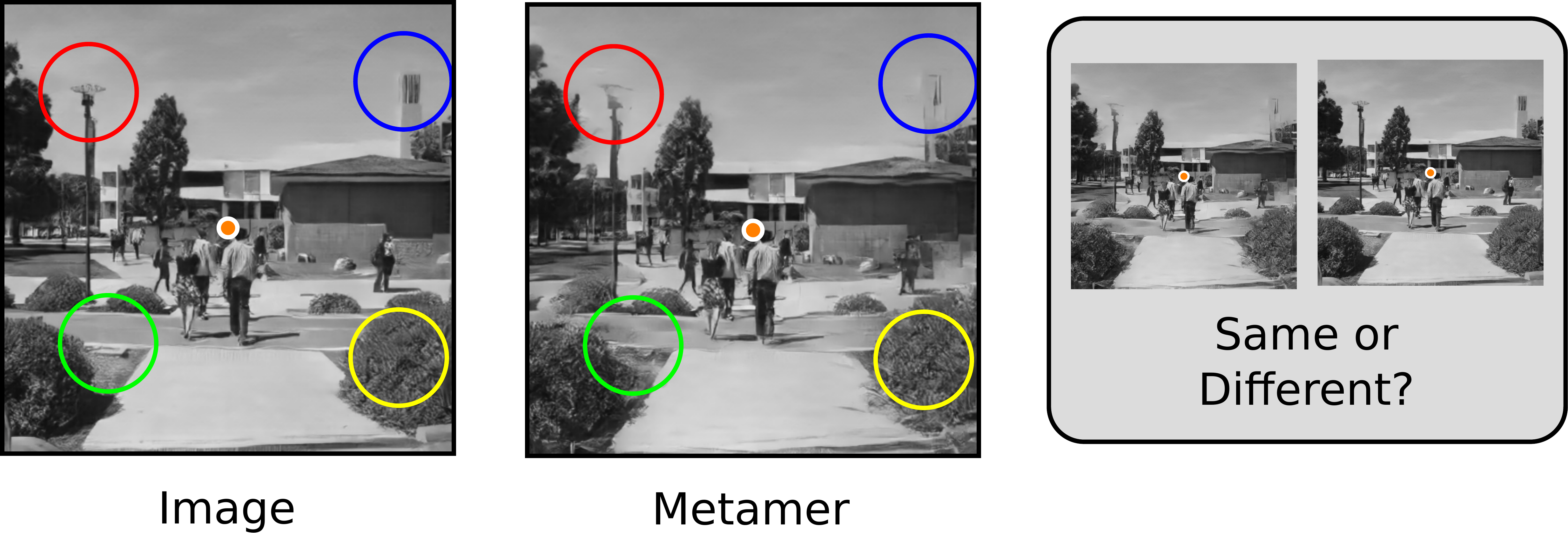}
  \caption{Two visual metamers are physically different images that when fixated on the orange dot (center), 
  should remain perceptually indistinguishable to each other for an observer. Colored circles highlight different distortions
  in the visual field that observers do not perceive in our model.
  }\label{fig:Metamer_Cartoon}
\end{figure}

\vspace{-10pt}
The history of metamers originally started through color matching theory, where two light sources were used to match 
a test light's wavelength, until both light sources
are indistinguishable from each other producing what is called a \textit{color metamer}. 
This leads to the definition of visual metamerism: 
when two physically different
stimuli produce the same perceptual response (See Figure~\ref{fig:Metamer_Cartoon} for an example). 
Motivated by~\cite{balas2009summary}'s work of local texture matching in the periphery as a mechanism that 
explains visual crowding,~\cite{freeman2011metamers} were the first to create such point-of-fixation driven metamers
through such local texture matching models that tile the entire 
visual field given log-polar pooling regions that simulate the V1 and V2 receptive field sizes,
as well as having global image statistics that match the metamer with the original 
image. 
The essence of their algorithm is to use gradient descent
to match the local texture~(\cite{portilla2000parametric}) and image statistics of 
the original image throughout the visual field given a 
point of fixation until convergence thus producing two images that are perceptually indistinguishable to each other.

However, metamerism research currently faces 2 main limitations:
The first is that metamer rendering faces no unique solution. 
Consider the potentially trivial examples of having an image $I$ and its 
metamer $M$ where all pixel values are identical except for one which is
set to zero (making this difference unnoticeable), or the case where the metameric response arises from 
an imperceptible equal 
perturbation across all pixels as suggested in ~\cite{johnson2016perceptual,freeman2011metamers}.
This is a concept similar to Just Noticeable Differences~(\cite{lubin1997human,daly1992visible}).
However, like the work of \cite{freeman2011metamers,keshvari2016pooling,rosenholtz2012summary,balas2009summary}, 
we are interested in creating point-of-fixation driven metamers, which
create images that preserve information in the fovea, yet lose spatial 
information in the periphery such that this loss is unnoticeable contingent of a point of 
fixation (Figure~\ref{fig:Metamer_Cartoon}). 
The second issue is that the current state of the art for a full field of view rendering of a $512\text{px}\times512$px 
metamer takes 6 hours for a grayscale image and roughly a day for a color image. 
This computational constraint makes data-driven 
experiments intractable if they require thousands of metamers.
From a practical perspective, creating metamers that are quick to compute may lead to 
computational efficiency in rendering of VR foveated displays
and creation of novel neuroscience experiments that require metameric stimuli such as gaze-contingent displays, or 
metameric videos for fMRI, EEG, or Eye-Tracking.

We think there is a way to capitalize metamer understanding and rendering given the developments made in the field of 
\textit{style transfer}.
We know that the original model of~\cite{freeman2011metamers} consists of a local texture matching procedure 
for multiple pooling regions
in the visual field as well as global image content matching. If we can find a way to perform localized style transfer
with proper texture statistics for all the pooling regions in the visual field, 
and if the metamerism via texture-matching 
hypothesis is correct -- we can in theory successfully render a metamer.

Within the context of style transfer, we would want a complete and flexible framework where a \emph{single} 
network can encode \emph{any} style (or texture) without the need 
to re-train, and with the power
of producing style transfer with a single forward pass, thus enabling real-time applications. 
Furthermore, we would want such framework to also
control for spatial and scale factors~(\cite{gatys2016controlling}) to enable foveated pooling~(\cite{akbas2017object,deza2016can})
which is critical in metamer rendering. 
The very recent work of \cite{huang2017adain}, provides such framework through adaptive instance normalization~(AdaIN), where 
the content image is stylized by adjusting the mean and standard deviation 
of the channel activations of the encoded representation to match with the style. 
They achieve results that rival those 
of~\cite{ulyanov2016texture,johnson2016perceptual}, with the added benefit of not being limited to a single texture in a feed-forward pipeline.

In our model: we stack a peripheral architecture 
on top of a VGGNet~(\cite{simonyan2014very}) in its encoded feature space, to map an image into a perceptual space.
We then add internal noise in the encoded space of our model as a characterization
that perceptual systems are noisy. We find that inverting such modified image representation via a decoder 
results in a metamer. This breaks down our model into 
a foveated feed-forward `auto' style transfer network, where the input image plays the role both 
of the content and the style, and internal network noise (stylized with the content statistics) 
serves as a proxy for intrinsic image texture. 
While our model uses AdaIN for style transfer and a VGGNet for texture statistics,
our pipeline is extendible to other models that successfully execute style transfer and 
capture proper texture statistics~(\cite{ustyuzhaninov2017texture}).


\begin{figure}[t]
\centering
\includegraphics[width=1.0\columnwidth,clip=true,draft=false,]{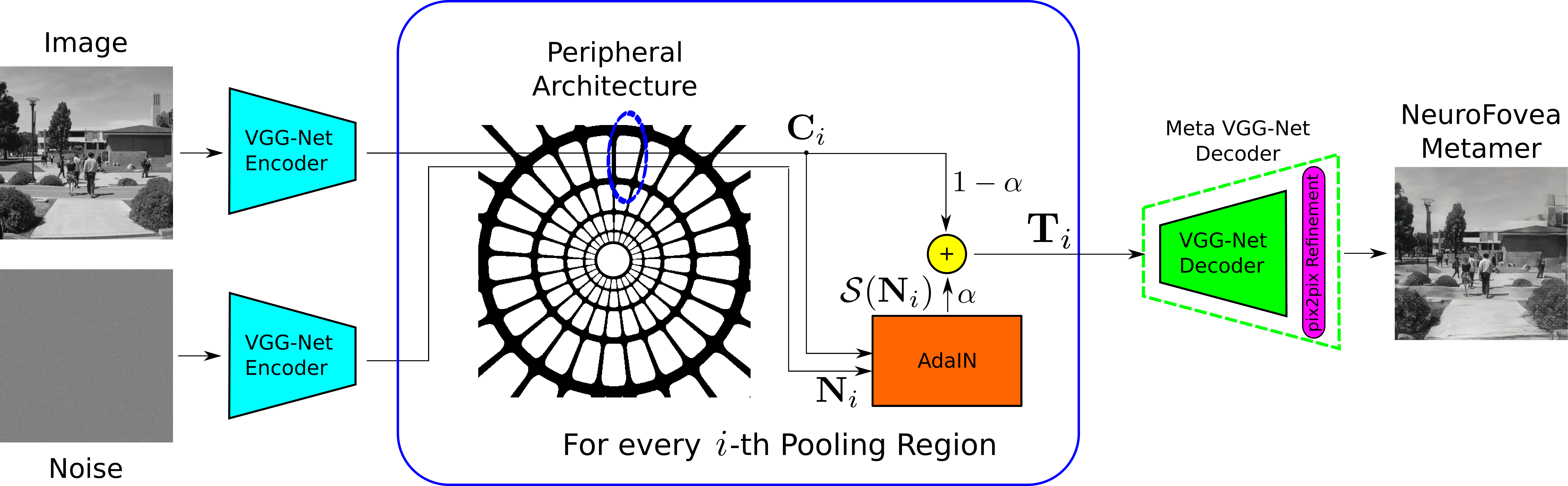}
  \caption{The NeuroFovea metamer generation schematic: 
  An input image and a noise patch are fed through a VGG-Net encoder into a new feature space. 
  Through spatial control we can produce an interpolation for each pooling region in such 
  feature space between the stylized-noise (texture), and the content (the input image). 
  This is how we successfully impose both global image
  and local texture-like constraints in every pooling region. 
  The metamer is the output of the pooled (and interpolated) feature vector
  through the Meta VGG-Net Decoder. 
  }\label{fig:Model_Diagram}
\end{figure}




\vspace{-10pt}
\section{Design of the NeuroFovea model}
\vspace{-10pt}
To construct our metamer we propose the following statement:
A metamer $M$ can be rendered by transferring $k$ \emph{localized} styles over a content image $I$, controlled by a set of 
style-to-content ratios $\alpha_i$ for every pooling region ($i$-th receptive field). 
More formally, our goal is to find a Metamer function $\mathbf{M}(\circ):I\rightarrow M$, 
where an input image $I\in\mathbb{R}^L$ is fed through a 
VGG-Net encoder $\mathcal{E}(\cdot):\mathbb{R}^{L}\rightarrow \mathbb{R}^{D}$ 
which is both the content and the style image, 
  to produce the content feature $\mathbf{C}\in\mathbb{R}^D$, 
  where $\mathbf{C}=\mathcal{E}(I)$ as shown in Figure~\ref{fig:Model_Diagram}. 
  Let $L=C\times H\times W$, and $D=C'\times H' \times W'$
  where $\{C,C'\},\{H,H'\},\{W,W'\}$ are the image/layer channels, height, width given the convolutional structure of the encoder (we drop fully connected layers).
  A noise patch colored via ZCA~(\cite{bell1995information}) to match the content image's mean and variance 
  $\mathcal{N}\sim(\mu_I,\sigma_I^2)\in\mathbb{R}^L$ is also fed 
  through the same VGG-Net encoder producing the
  noise feature $\mathbf{N}\in\mathbb{R}^D$, where $\mathbf{N}=\mathcal{E}(\mathcal{N})$. 
  This is the internal perceptual noise of the system which will later on serve us as a proxy for texture encoding.
  These vectors are masked through spatial control~\textit{a la}~\cite{gatys2016controlling}, 
  and the noise is stylized via $\mathcal{S}(\cdot):\mathbb{R}^D\rightarrow \mathbb{R}^D$ with 
  the content which encodes the texture representation of the content in the feature space through Adaptive Instance Normalization (AdaIN).
  A target feature $\mathbf{T}_i\in\mathbb{R}^D$ is defined as an 
  interpolation between the stylized noise $\mathcal{S}(\mathbf{N}_i)$ and the content $\mathbf{C}_i$ modulated by $\alpha$, in 
  the feature space $\mathbb{R}^D$ for every $i$-th pooling region:
\begin{equation}
\label{Eq:Homotopy}
\mathbf{T}_{i}(I|\mathcal{N};\alpha) = (1-\alpha)\mathbf{C}_i(I) + \alpha \mathcal{S}(\mathbf{N}_i)
\end{equation}
\begin{figure}[b]
\centering
\includegraphics[width=1.0\columnwidth,clip=true,draft=false,]{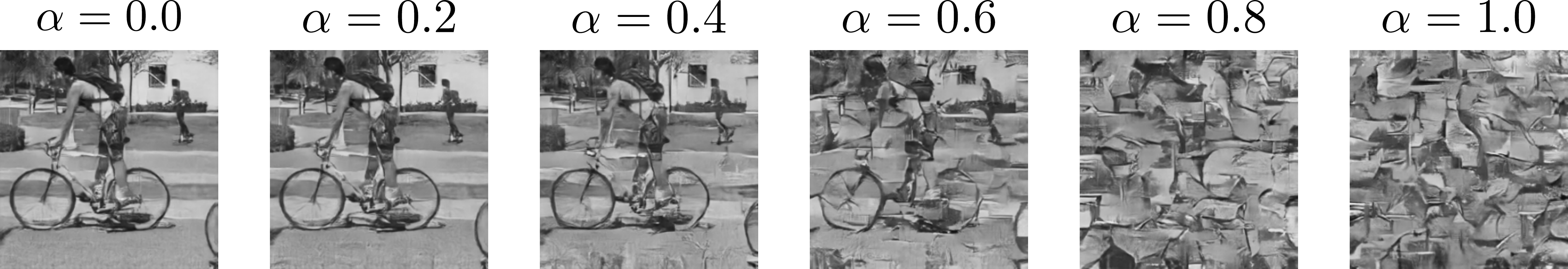}
\vspace{-10pt}
  \caption{Interpolating between an image's intrinsic content and texture via a convex combination in the output of 
  the VGG19 Encoder $\mathcal{E}$. Here 
  we are treating the patch as a single pooling region. In our model, this interpolation given Eq.~\ref{Eq:Homotopy} 
  is done for every pooling region in the visual field.
  }\label{fig:Homotopy_Demo}
\vspace{-10pt}
\end{figure}
  In other words, in our quest to probe for metamerism, we are finding an intermediate representation 
  (the convex combination) between two 
  vectors representing the image and its texturized version (the stylized noise)
  in $\mathbb{R}^D$ per pooling region as seen in Figure~\ref{fig:Homotopy_Demo}. 
  Within the framework of style transfer, we could think of this as a 
  content-vs-style or structure-vs-texture
  tradeoff, since the style and the content image are the same.
  Similar interpolations have been explored in~\cite{henaff2015geodesics} via a joint pixel and network space minimization.
  The final target feature vector
  $\mathbf{T}$ is the masked sum of every $\mathbf{T}_i$ with spatial control masks $w_i$ 
  s.t. $\mathbf{T}=\sum{w_i \mathbf{T}_i}$.
  The metamer is the output of the Meta VGG-Net decoder $\mathcal{D}(\cdot)$ on $\mathbf{T}$, 
  where the decoder receives only \emph{one} 
  vector $(\mathbf{T})$ and produces a global decoded output.
Our Meta VGG-Net Decoder compensates for small 
artifacts by stacking a \textit{pix2pix}~\cite{isola2016image} U-Net refinement module 
which was trained on the Encoder-Decoder outputs to map to the original 
high resolution image. Figure~\ref{fig:Model_Diagram} fully 
describes our model, and the metamer transform is computed via: 
\vspace{-5pt}
\begin{equation}
\label{Eq:Metamer}
\mathbf{M}(I|\mathcal{N};\bar{\alpha}) = \mathcal{D}(\mathcal{E}_{\sum}(I|\mathcal{N};\bar{\alpha})) = \mathcal{D}(\sum_{i=1}^k w_i \big[ (1-\alpha_i)\mathcal{E}_i(I) + \alpha_i\mathcal{S}(\mathcal{E}_i(\mathcal{N}))  \big])
\end{equation}
where $\mathcal{E}_{\sum}$ is the foveated encoder that is defined as the sum of encoder outputs over all 
the $k$ pooling regions (our spatial controls masks $w_i$) in the 
visual field. 
Note that the decoder
was not trained to generate metamers, but rather to invert the encoded image and act as $\mathcal{E}^{-1}$. It happens to be the case
that perturbing the encoded representation in the direction of the stylized noise by an amount specified by the size of the pooling regions,
outputs a metamer.
Additional specifications and training of our model can be seen in the Supplementary Material.
  
  \subsection{Model Interpretability}
\begin{wrapfigure}{r}{0.35\textwidth}
\centering
\includegraphics[width=0.35\columnwidth,clip=false,draft=false,]{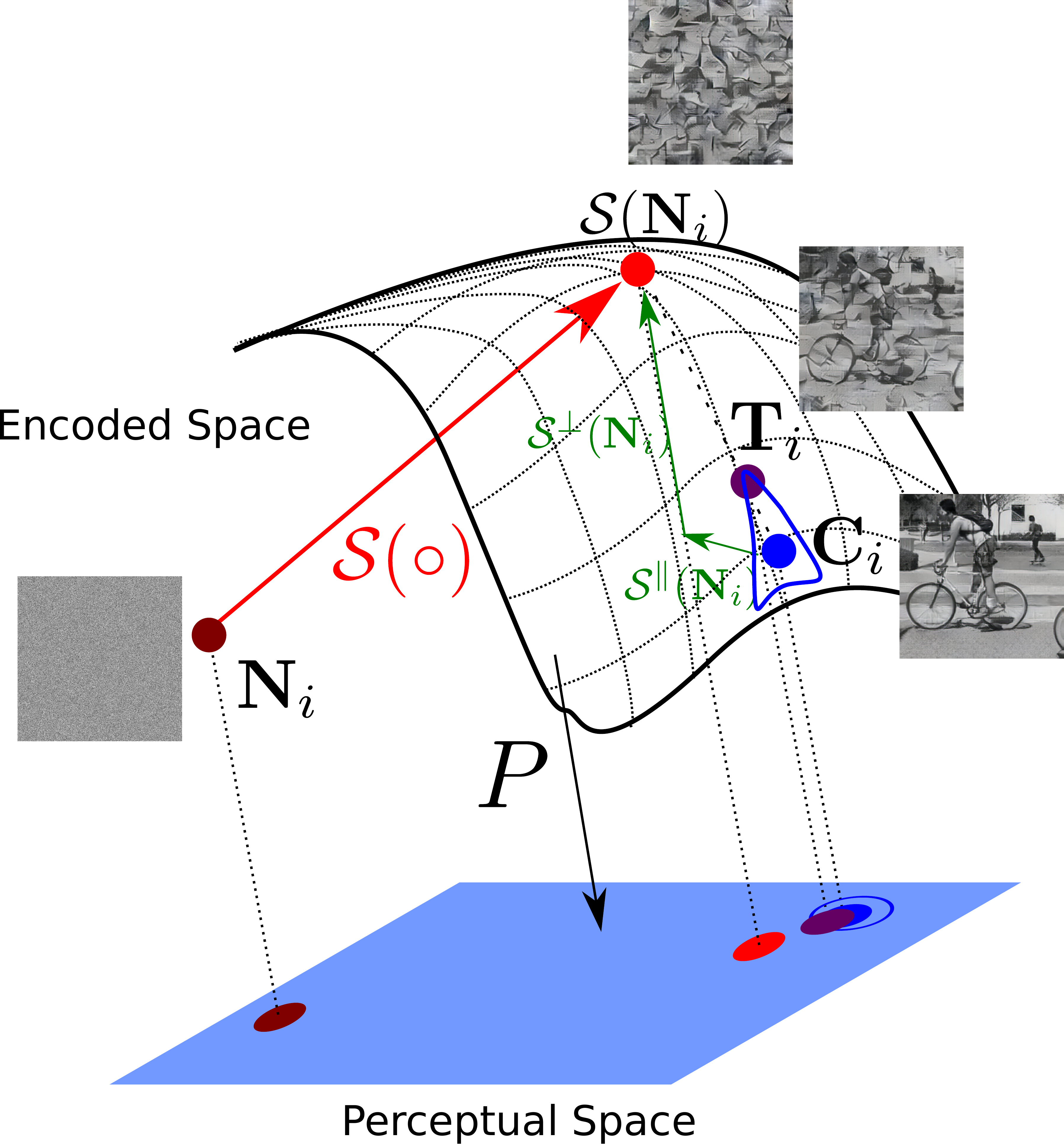}
\vspace{-10pt}
  \caption{Perceptual Projection.
  }\label{fig:Perceptual_Space}
\vspace{-15pt}
\end{wrapfigure}

  Within the framework of metamerism where distortions lie on the perceptual null space as proposed 
  initially in color matching theory, and
  also in~\cite{freeman2011metamers} for images, 
  we can think of our model as a direct transform that
  is maximizing how much information to discard depending on the texture-like properties of the image and the size of the receptive 
  fields. Consider the following: if our interpolation is projected from the encoded space to the perceptual space via $P$,
  from Eq.~\ref{Eq:Homotopy} we get
$ P\mathbf{T}_{i} = P(1-\alpha)\mathbf{C}_i(I) + P(\alpha) \mathcal{S}(\mathbf{N}_i)$, it follows that for each receptive field:
\begin{equation}
P\underbrace{\mathbf{T}_{i}}_\text{metamer} = P\underbrace{\mathbf{C}_i}_\text{image} + P\underbrace{\alpha(\mathcal{S}^{\bot}(\mathbf{N}_i)+\mathcal{S}^{\parallel}(\mathbf{N}_i))}_\text{distortion}
\end{equation}
by decomposing $\mathcal{S}(\mathbf{N}_i)-\mathbf{C}_i=\mathcal{S}^{\bot}(\mathbf{N}_i)+\mathcal{S}^{\parallel}(\mathbf{N}_i)$,
where $\mathcal{S}^{\parallel}$ is the projection of the difference vector on the perceptual space,
and $\mathcal{S}^{\bot}(\mathbf{N}_i)$
is the orthogonal component perpendicular to such vector which lies in the perceptual null space 
$(P\mathcal{S}^{\bot}(\mathbf{N}_i)=\vec{0})$. 
The value of these components will change depending on the location of $\mathbf{C}_i$ and $\mathcal{S}(\mathbf{N}_i)$,
and the geometry of the encoded space.
If $||\mathcal{S}^{\parallel}(\mathbf{N}_i)||_2^2<\eps$, (\text{i.e.}
the image patch has strong texture-like properties),
then $\alpha$ can vary above its critical value given that $\mathcal{S}^{\bot}(\mathbf{N}_i)$ is in the null space of $P$ and the 
distortion term will still be small;
but if $||\mathcal{S}^{\parallel}(\mathbf{N}_i)||_2^2>\eps$, $\alpha$ can not exceed its critical value
for the metamerism condition to hold $(P\mathbf{T}_{i}\approx P\mathbf{C}_i)$.
Thus our interest is in computing the maximal \textit{average} 
amount of distortion (driven by $\alpha$) given human sensitivity before observers can tell the 
difference. This is illustrated in Figure~\ref{fig:Perceptual_Space} via the blue circle around $\mathbf{C}_i$ in the perceptual space 
which shows the \textit{metameric boundary} for any distortion.

One can also see the resemblance of the model to a noisy communication system in the context of information theory.
The information source is the image $I$, 
the transmitter and the receiver are the encoder and decoders $(\mathcal{E},\mathcal{D})$ respectively,
and the noise source is the encoded noise patch $\mathcal{E}(\mathcal{N})$ imposing texture 
distortions in the visual field, and the destination is the metamer $M$.
Highlighting this equivalence
is important as metamerism can also be explored within the context of image compression and rate-distortion theory as 
in~\cite{balle2016end}. Such approaches are 
beyond the scope of this paper, however they are worth exploring in future work as most metamer models
purely involve texture and image analysis-synthesis matching paradigms that are gradient-descent based.

\vspace{-10pt}
\section{Hyperparameteric nature of our model}
\vspace{-10pt}
Similar to our model, the~\cite{freeman2011metamers} model (hereto be abbreviated FS)
requires a scale parameter $s$ which controls the rate of growth of the receptive fields 
as a function of eccentricity. 
This parameter should be maximized such that an upperbound for 
perceptual discrimination is found. 
Given that texture and image matching occurs in each one of the pooling regions: 
a high scaling factor will likely make the image
rapidly distinguishable from the original as distortions are more apparent in the periphery. Conversely,
a low scaling factor might gaurantee metamerism even if the texture statistics are not fully correct given that 
smaller pooling regions will simulate weak effects of crowding. 
Low scaling factors in that sense are potentially uninteresting -- 
it is the value up until humans can tell the difference that is critical~(\cite{lubin1997human}).
FS set out to find such critical value via a psychophysical experiment where they 
perform the following single-variable optimization to find such upper bound:
\begin{equation}
\label{Eq:Optim_Main_FS}
 s_0 = \argmax_{s} ~ \mathbb{E}[d'(s|\theta_{obs})] 
\end{equation}
s.t. $ 0 < d'(s|\theta_{obs}) < \epsilon$, where 
$d'=\Phi^{-1}(\text{HR})-\Phi^{-1}(\text{FA})$ is the index of detectability for each observer $\theta_{obs}$, $\Phi$ is
the cumulative of the gaussian distribution, and $\text{HR}$ and $\text{FA}$ are the hit rate and false alarm rates
as defined in~\cite{green1966signal}.
However, our model is different in regards to a set of hyperparameters
$\bar{\alpha}$ that we must estimate everywhere in the visual field as summarized by the $\gamma$ function, where
we assume $\alpha$ to be tangentially isotropic:
\begin{equation}
\alpha = \gamma(\circ;s)
\end{equation}
where each $\alpha$ represents the maximum amount of distortion (Eq.~\ref{Eq:Homotopy}) 
that is allowed for every receptive field 
in the visual periphery before an observer will notice.
At a first glance, it is not trivial to know if $\alpha$ should be a function
of scale, retinal eccentricity, receptive field size, image content or potentially a combination of the before-mentioned (hence 
the $\circ$ in the $\gamma$ function's argument).

\begin{figure}[!t]
\centering
\includegraphics[width=1.0\columnwidth,clip=true,draft=false,]{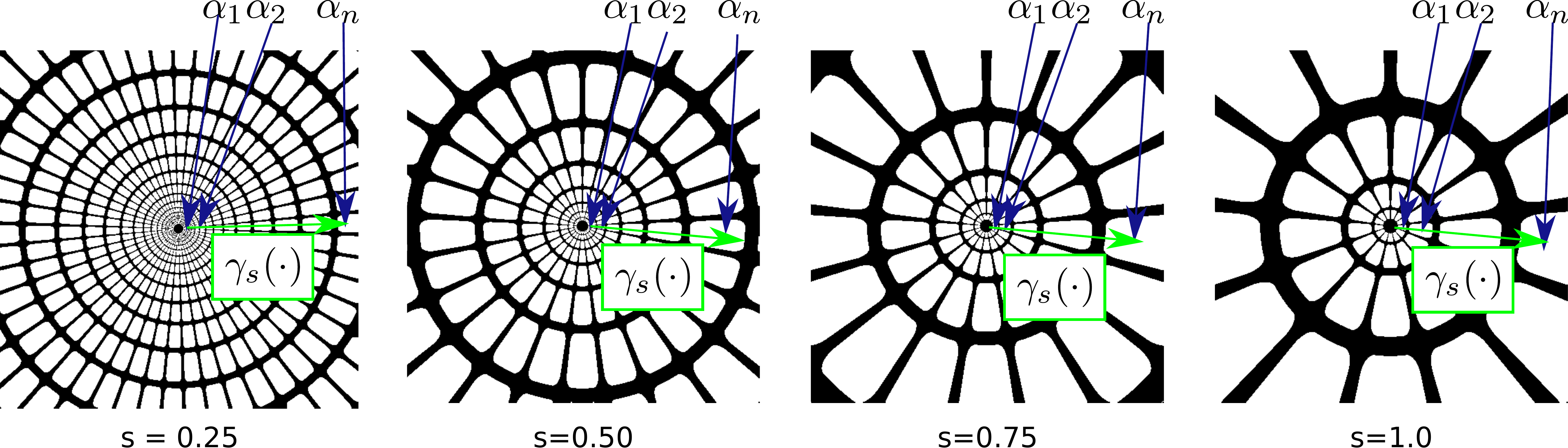}
  \caption{Potential issues of 
  psychophysical intractability for the joint estimation of $(s)$ and 
  $\gamma(\cdot)$ as described by our model. Running a psychophysical experiment that runs an exhaustive search for 
  upper bounds for the scale and distortion parameters for every receptive field
  is intractable. The goal of Experiment 1
  is to solve this intractabitilty posed formally in Eq.~\ref{Eq:Optim_Main} via a simulated experiment.
  }\label{fig:Metamers_Scale}
\end{figure}


Thus, the motivation of $\alpha$ seems uncertain and perhaps un-necessary from the Occam's razor perspective of model simplicity.
This raises the question: Why does the 
FS model not require any additional hyperparameters, requiring only a single scale $(s)$ parameter? 
The answer lies in the nature of their model which is gradient descent based and where local texture 
statistics are matched for every pooling region 
in the visual field, while preserving global image structural information. When such condition is reached, no further synthesis steps are required
as it is an equilibrium point. 
Indeed, the experiments of ~\cite{wallis2016testing} have shown that images do not remain metameric
if the structural information of a pooling region is discarded while purely retaining the 
texture statistics of~\cite{portilla2000parametric}. This motivates the purpose of $\alpha$ 
where we interpolate between structural and texture representation.
Thus our goal is to find that equilibirum point in one-shot, 
given that our model is purely feed-forward and requires no gradient-descent (Eq.~\ref{Eq:Metamer}).
At the expense of this artifice, we run into the challenge of facing
a multi-variable optimization problem that has the risk of being psychophysically intractable. Analogous to FS,
we must solve:
\begin{equation}
\label{Eq:Optim_Main}
 s_0,\bar{\alpha_0} = \argmax_{s,\bar{\alpha}} ~ \mathbb{E}[d'(s,\bar{\alpha}|\theta_{obs})]
\end{equation}
s.t. $ 0 <  d'(s,\bar{\alpha}|\theta_{obs}) < \epsilon$.
Figure~\ref{fig:Metamers_Scale} shows the potential intractability:
each observer would have to run multiple rounds of an ABX experiment for a collection of many scales and $\alpha$ values for each location
in the visual field. Consider: ($S$ scales) $\times$ ($k$ pooling regions) $\times$ ($\alpha_m$ step size for each $\alpha$) 
$\times$ ($N$ images) $\times$ ($w$ trials): $SkN\alpha_m w$ trials per observer.

We will show in Experiment 1 that one solution to Eq.~\ref{Eq:Optim_Main} is to find a relationship between each set of $\alpha$'s and the scale, expressed via the $\gamma$ 
function. This requires a two stage 
process: 1) Showing that such $\gamma$ exists; 2) Estimate $\gamma$ given $s$. 
If this is achieved, we can
relax the multi-variable optimization into a single variable optimization problem, 
where $ 0 <  d'(s,\gamma(\circ;s)|\theta_{obs}) < \epsilon$, and:
\begin{equation}
\label{Eq:Optim_Relief}
 s_0 = \argmax_{s} ~ \mathbb{E}[d'(s,\gamma(\circ;s)|\theta_{obs})]
\end{equation}
\vspace{-20pt}
\section{Experiments}
\begin{figure}[!t]
\centering
\includegraphics[width=1.0\columnwidth,clip=true,draft=false,]{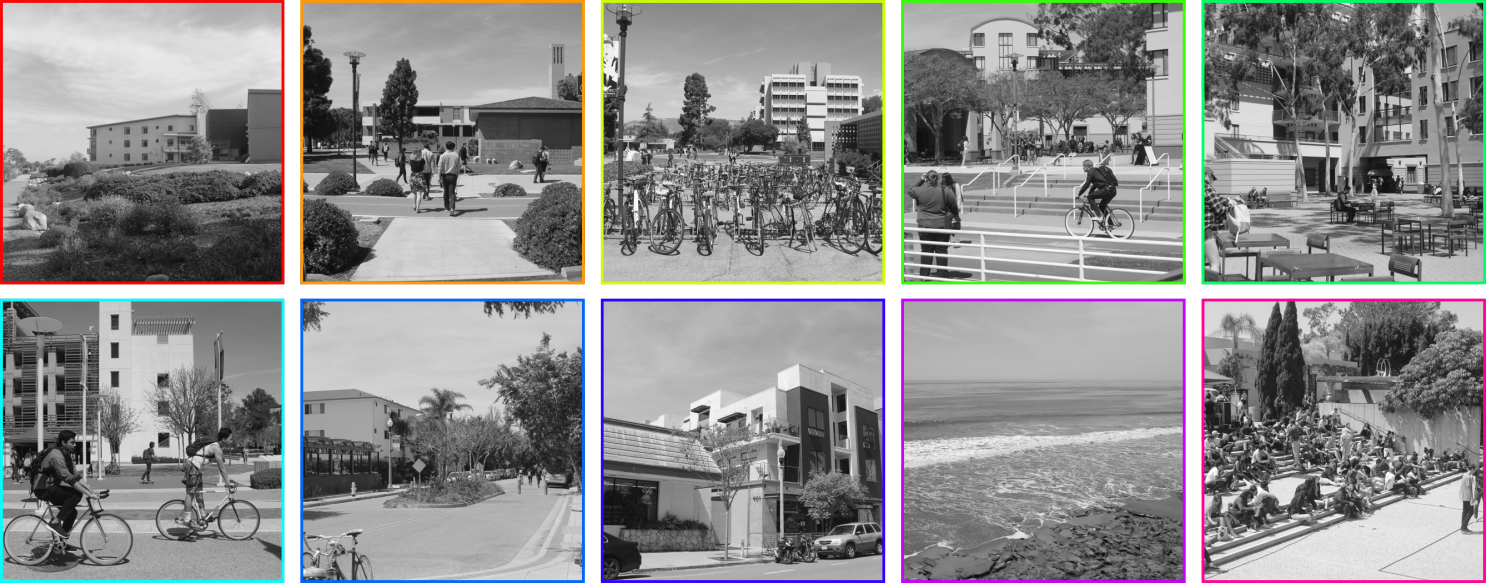}
\caption{A color-coded collection of images used in our experiments. 
}
\label{fig:Images}
\end{figure}

\vspace{-10pt}
The goal of Experiment 1 is to estimate $\gamma$ as a function of $s$  via a computational simulation as a proxy for 
running human psychophysics. Once it is computed, we have reduced our minimization to a tractable single 
variable optimization problem. We will then proceed to Experiment 2 where
we will perform an ABX experiment on human observers by varying the scale to render visual metamers as 
originally proposed by FS. We will use the images shown in Figure~\ref{fig:Images} for both our experiments.

\subsection{Experiment 1: Estimation of model hyperparameters via perceptual optimization}
\label{sec:Experiment1}
\textbf{Existence and shape of $\gamma$:}
Given some biological priors, we would like $\gamma$ to satisfy these properties:
\begin{enumerate}
 \item $\gamma: Z \rightarrow \alpha$ s.t. $Z\in[0,\infty),\alpha\subset[0,1)$, where $z\in Z$ is parametrized by the size (radius) of each 
 receptive field (pooling region) which grows with eccentricity in humans. 
 \item $\gamma$ is continuous and monotonically non-decreasing since more information should not be gained given 
 larger crowding effects as receptive field size increases in the periphery. 
 \item $\gamma$ has a unique zero at $\gamma(0) = 0$. Under ideal assumptions there is no loss of information in the fovea, where the size of the receptive
 fields asymptotes to zero.
\end{enumerate}

Indeed, we found that $\gamma$ is sigmoidal, and is a function of $z$, parametrized by $s$:
\begin{equation}
\label{eq:sigmoide}
 \gamma(z;s) = a+\frac{b}{c+\text{exp}(-dz)} = -1+\frac{2}{1+\text{exp}(-d(s)z)}
\end{equation}


\begin{wrapfigure}{r}{0.5\textwidth}
\centering
\includegraphics[width=0.5\columnwidth,clip=false,draft=false,]{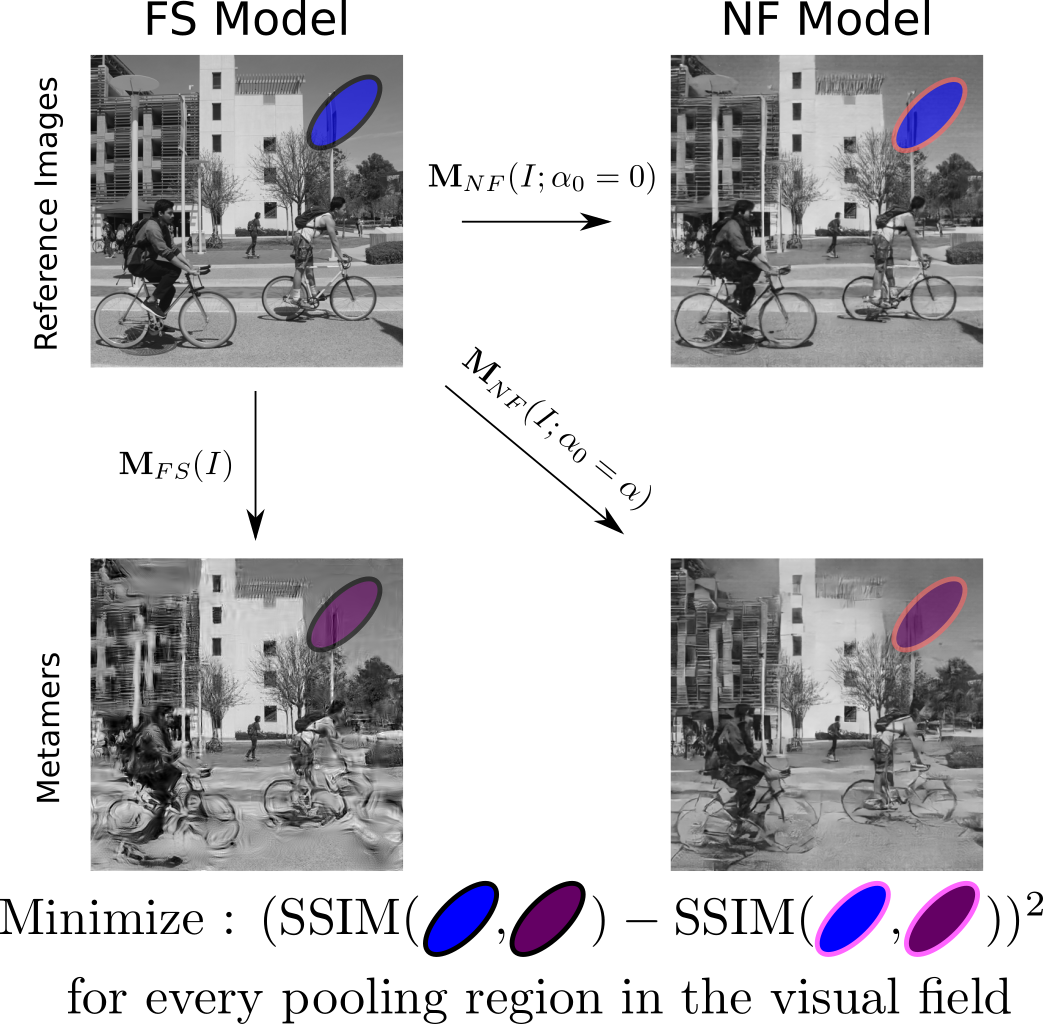}
\vspace{-10pt}
  \caption{Perceptual optimization.
  }\label{fig:Perceptual_Optimization}
\vspace{-10pt}
\end{wrapfigure}

\textbf{Estimation of $\gamma$:}
To numerically estimate the amount of $\alpha$-noise 
distortion for each receptive field in our metamer model we need to find a way to 
simulate the perceptual loss made by a human observer when trying to discriminate 
between metamers and original images.
We will define a perceptual loss $\mathcal{L}$ that 
has the goal of matching the distortions via SSIM of a gradient descent based method such as 
the FS metamers, and the NeuroFovea metamers (NF) with their reference images -- a strategy similar 
to~\cite{laparra2017perceptually} used for perceptual rendering.
We chose SSIM as it is a standard IQA metric that is monotonic with human judgements, although other metrics such as MS-SSIM and
IW-SSIM show similar tuning properties for $\gamma$ as shown in the Supplementary Material.
Indeed the \textit{reference} image $I'$ for the NF metamer is limited by the autoencoder-like nature of the model
where the bottleneck usually limits perfect reconstruction s.t. $I'= \mathcal{D}(\mathcal{E}(I))|_{(\alpha=0)}$, 
where $I'\rightarrow I$, and they are only equal
if the encoder-decoder pair $(\mathcal{E},\mathcal{D})$ allows for lossless compression. 
Since we can not define a direct loss function $\mathcal{L}$ between the metamers, 
we will need their reference images to define a convex surrogate loss function $\mathcal{L}_R$. 
The goal of this function should be to match the perceptual loss of both metamers \textit{for each receptive field} $k$ 
when compared to their reference images: 
the original image $I$ for 
the FS model, and the decoded image $I'$ for the NF model: 
\vspace{-10pt}
\begin{equation} \label{eq:surrogate}
\mathcal{L}_R(\alpha|k) = \mathbb{E}(\Delta\text{-SSIM})^2 = \frac{1}{N} \sum_{j=1}^N (\text{SSIM}(M_{FS}^{(j,k)},I^{(j,k)})-\text{SSIM}(M_{NF}^{(j,k)}(\gamma_s),I'^{(j,k)}))^2
\end{equation}
and $\alpha_i$ should be minimized for each $k$ pooling region via: $\alpha_0 = \argmin_{\alpha} \mathcal{L}_R(\alpha|k)$ for the collection of $N$ images.
The intuition behind this procedure is shown in Figure~\ref{fig:Perceptual_Optimization}. Note that if
$I'=I$,~\textit{i.e.} there is perfect lossless compression and reconstruction given the choice of encoder and decoder, 
then the optimization is performed with reference to the same original image. This is an important observation as the 
reconstruction
capacity of our decoder is limited despite $\mathbb{E}(\text{MS-SSIM}(I,I')=0.86\pm0.04$. 
Only using the original image in the optimization yields poor local minima at $\alpha=0$.
Despite such limitation, we show that reference metamers can still be achieved for our 
lossy compression model.

\begin{figure}[!t]
\centering
\includegraphics[width=1.0\columnwidth,clip=true,draft=false,]{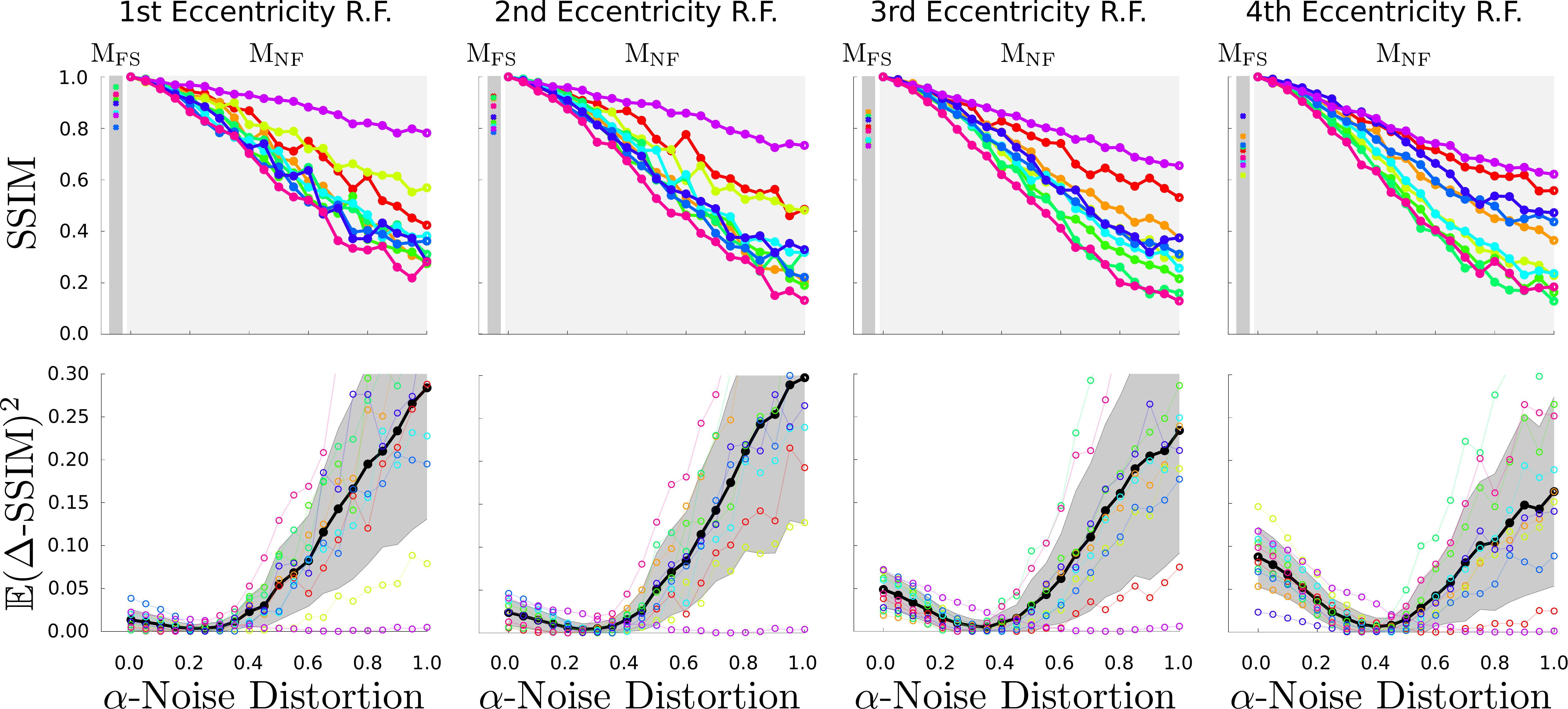}
\vspace{-10pt}
  \caption{The result of each SSIM (top) for Experiment 1 for a scale of $s=0.3$
  where we find the critical $\alpha$ for each receptive field ring as we 
  minimize $\mathbb{E}(\Delta\text{-SSIM})^2$ (bottom). $\mathbb{E}(\Delta\text{-SSIM})^2$ is minimized by matching the
  perceptual distortion of the~\cite{freeman2011metamers} ($M_{FS}$) and NeuroFovea ($M_{NF}$) metamers in Eq.~\ref{eq:surrogate}.
  Each color represents a different $512\times512$ image trajectory, the black line (bottom) shows the average. 
  Only the first 4 eccentricity dependent receptive fields are shown.
  }\label{fig:Metamers_Optimization}
\vspace{-10pt}
\end{figure}
\begin{figure}[!h]
\centering
\includegraphics[width=1.0\columnwidth,clip=true,draft=false,]{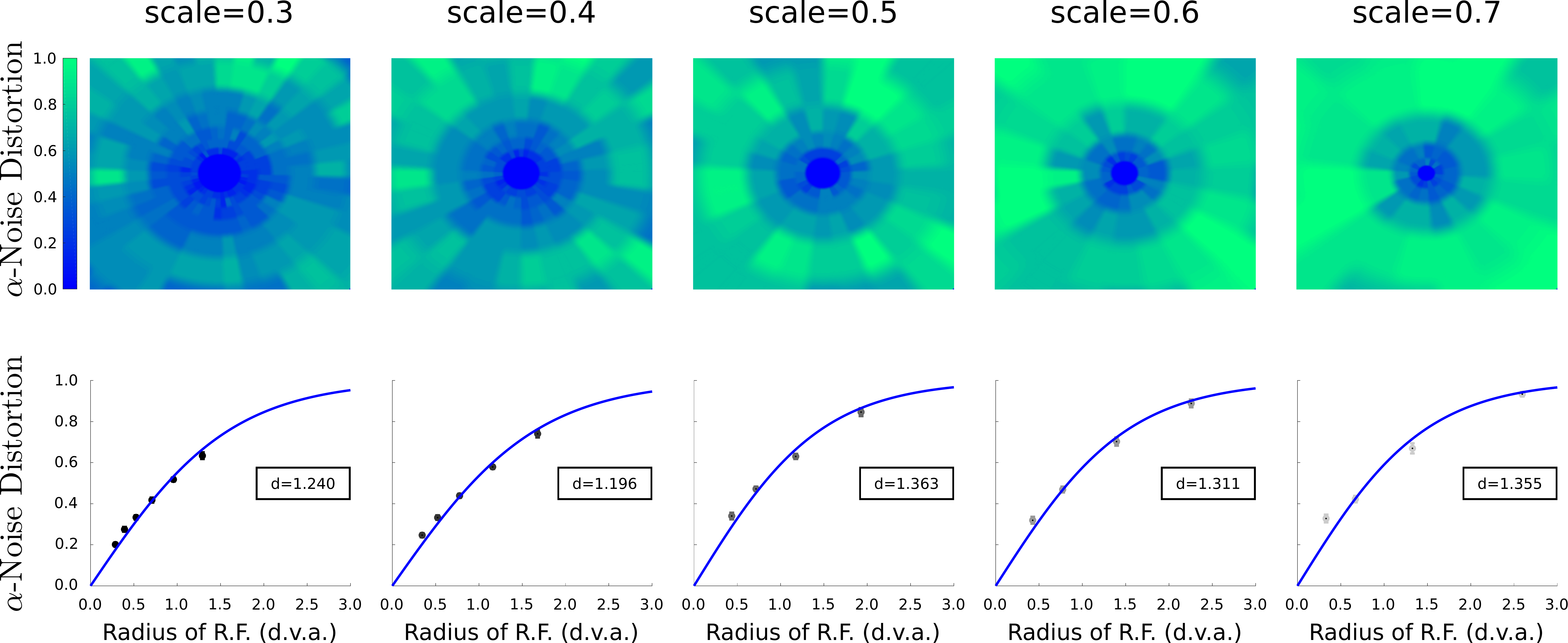}
\caption{Top: The average $\alpha$-noise distortion over the entire visual field for our 10 images
without assuming tangential homogeneity. Notice that on average, $\alpha$ increases radially. 
Bottom: The $\gamma(\cdot)$ which completely defines the $\alpha$-noise 
distortion for any receptive
field as a function of its size (radius).}
\label{fig:Alpha_Figure}
\end{figure}

\textbf{Results:} 
A collection of 10 images were used in our experiments.
We then computed the SSIM score for each FS and NF image paired with their reference image across each receptive field (R.F.) and averaged
those that belonged to the same retinal eccentricity.
Figure~\ref{fig:Metamers_Optimization} (top) shows these results, as well as the convex nature of the loss function displayed in the bottom. 
This procedure
was repeated for all the eccentricity-dependent receptive fields for a collection of 
5 values of scale: $\{0.3,0.4,0.5,0.6,0.7\}$. 
A sigmoid to estimate $\gamma$ was then fitted to each $\alpha$ per R.F. parametrized by scale via least squares. This gave us a collection of
$d$ values that control the slope rate of the sigmoid (Eq.~\ref{eq:sigmoide}).
These were $d:\{1.240,1.196,1.363,1.311,1.355\}$ respectively per scale, and $\{d\}=1.281$ for the ensemble of all scales.
We then conducted a 10000 sample permutation test between the pair of $(z_s,\alpha_s)$ points per scale 
and the ensemble of points across all scales $(\{z\},\{\alpha\})$
that verified that their variation is statistically non-significant $(p\geq0.05)$. 
Figure~\ref{fig:Alpha_Figure} illustrates the results from such procedure.
We can conclude that the parameters of $\gamma$ do not vary as we vary scale. In other words, the $\alpha=\gamma(z)$ function is fixed,
and the scale parameter itself which controls receptive field size will implicitly modulate 
the maximum $\alpha$-noise distortion with a unique $\gamma$ function. If the scale factor is small, the maximum noise distortion
in the far periphery will be small and \textit{vice versa} if the scale is large. 
We should point out that Figure~\ref{fig:Alpha_Figure} might suggest that
the maximal noise distortion is contingent on image content as the scores are not uniform tangentially for the receptive fields
that lie on the same eccentricity ring. Indeed, we did simplify our model by computing an average and fitting the sigmoid. However,
computing an average should approximate the maximal distortion for the receptive field size on that eccentricity in 
the \textit{perceptual space} for the human observer \textit{i.e.} the metameric boundary. 
We elaborate more on this idea in the discussion section.


\begin{figure}[!h]
\centering
\subfigure[A scale invariant $\gamma(\circ)$]{
    \includegraphics[width=0.35\columnwidth,clip=true,draft=false,]{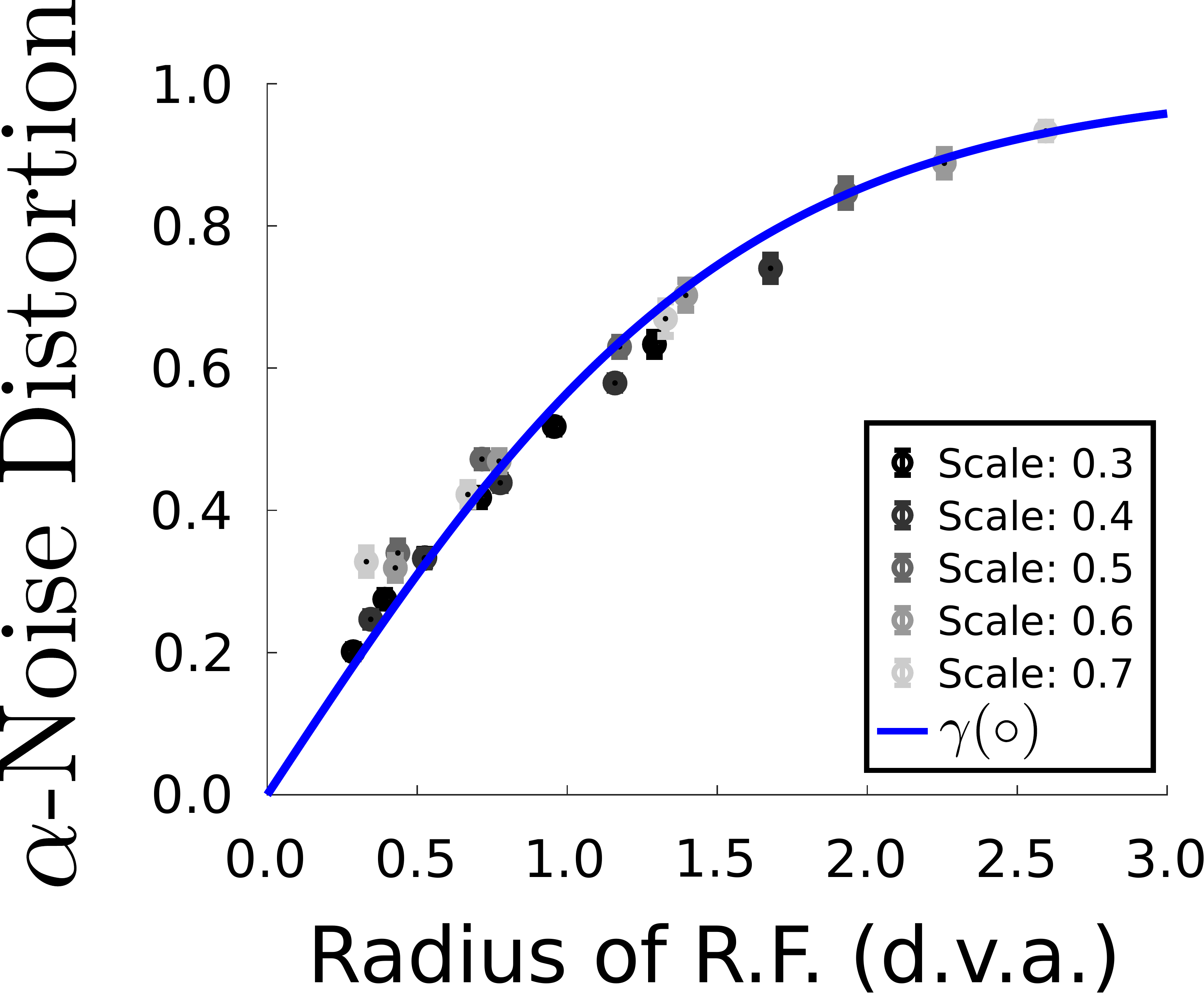}
    \label{fig:Gamma_Func}
}
\subfigure[Rendering metamers via varying $s$.]{
    \includegraphics[width=0.55\columnwidth,clip=true,draft=false,]{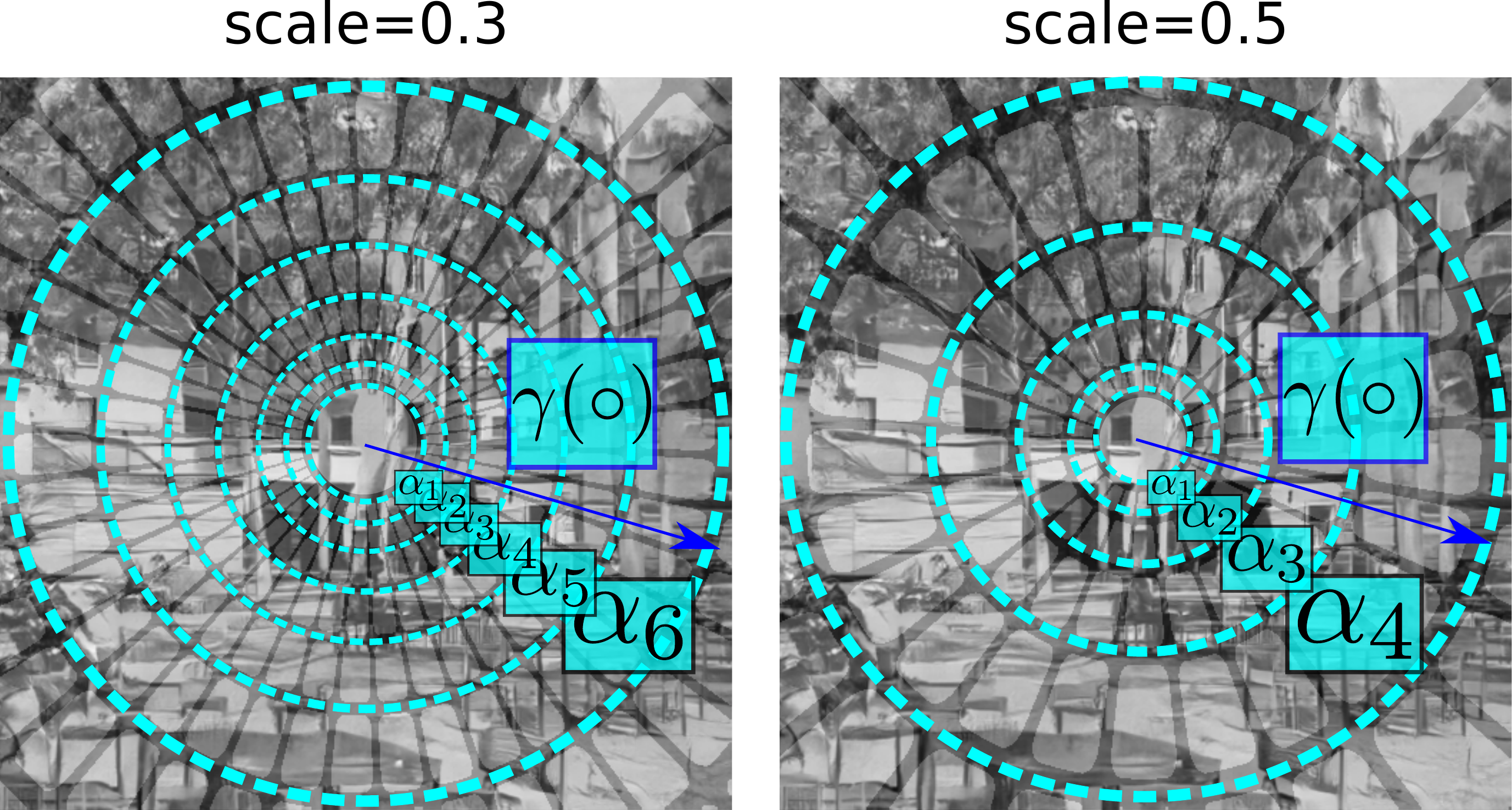}
    \label{fig:Variational}
}
\caption{Metamer generation proces for Experiment 2. We modulate
the distortion for each receptive field according to $\gamma$ 
to perform an optimization as in~\cite{freeman2011metamers}.
}\label{fig:Exp1_Results}
\vspace{-15pt}
\end{figure}

\vspace{-10pt}
\subsection{Experiment 2: Psychophysical Evaluation of Metamerism with human observers}
\begin{figure}[!t]
\centering
\includegraphics[width=1.0\columnwidth,clip=true,draft=false,]{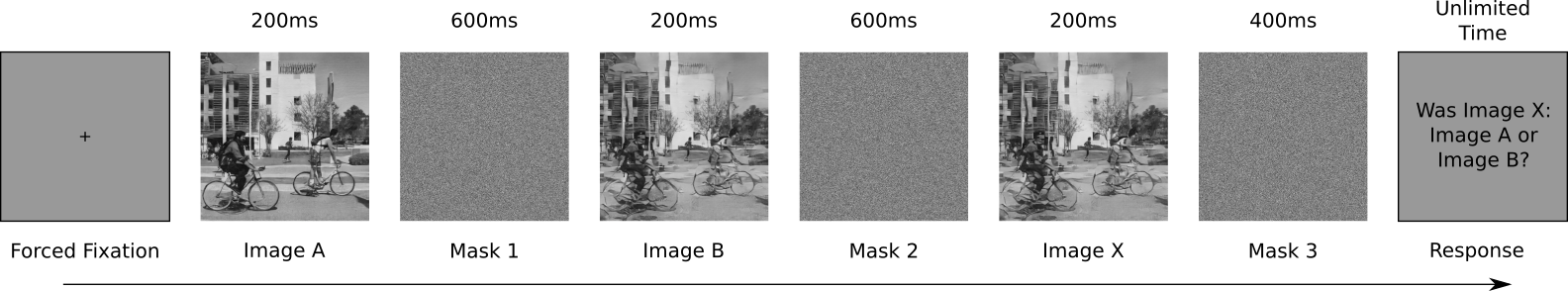}
\vspace{-10pt}
  \caption{Experiment 2 shows the ABX metamer discrimination task done by the observers. 
  Humans must fixate at the center of the image (no eye-movements) throughout the trial for it to be valid.
  }\label{fig:Experiment2}
\vspace{-10pt}
\end{figure}

\vspace{-10pt}
Given that we have estimated the value of $\alpha$ anywhere in the visual field via the $\gamma$ function, 
we can now render our metamers as a function
of the single scaling parameter $(s)$, as the receptive field size $z$ is also a function of $s$ as 
shown in Figure~\ref{fig:Exp1_Results}. The psychophysical optimization procedure
is now tractable on human observers and has the following form where $0 <  d'(s,\gamma(z(s);s)|\theta_{obs}) < \epsilon$:
\begin{equation}
\label{Eq:Optim_Final}
 s_0 = \argmax_{s} ~ \mathbb{E}[d'(s,\gamma(z(s))|\theta_{obs})]
\end{equation}
Inspired by the evaluations of~\cite{wallis2016testing}, 
we wanted to test our metamers on a group of observers performing two different ABX discrimination tasks in a roving design:
\begin{enumerate}
\item Discriminating between Synthesized images (Synth \textit{vs} Synth): 
This has been done in the original study of Freeman \& Simoncelli.
While this test does not gaurantee metamerism (Reference \textit{vs} Synth), 
it has become a standard evaluation when probing for metamerism.
\item Discriminating between the Synthesized and Reference images (Synth \textit{vs} Reference). 
This metamerism test, was not previously reported in~\cite{freeman2011metamers} for their original images
and is the most rigorous evaluation. Recently~\cite{Wallis378521} argued that any model that maps an image 
to white noise might gaurantee metamerism under the Synth~\textit{vs}~Synth condition but not against the original/reference image, 
thus is not a metamer.
\end{enumerate}
We had a group of 3 observers agnostic to the peripheral distortions
and purposes of the experiment performed an interleaved Synth vs Synth and Synth vs Reference experiment for 
NF metamers for the previous set of images~(Fig.~\ref{fig:Images}).
An SR EyeLink 1000 desk mount was used to monitor their gaze for the center forced fixation ABX task as 
shown in Figure~\ref{fig:Experiment2}. In each trial, observers were shown 3 images where their task
is to match the third image to the 1st or the 2nd. 
Each observer saw each of the 10 images 30 times per scaling factor (5) per discriminability type (2)
totalling 3000 trials per observer. 
Images were rendered at $512\times512$ px, and we fixed the monitor at 52cm viewing distance and $800\times600$px resolution 
so that the stimuli subtended $26\deg\times26\deg$. The monitor was linearly calibrated with a 
maximum luminance of $115.83\pm 2.12$ $cd/m^2$.
We then estimated the critical
scaling factor $s_0$, and absorbing factors $\beta_0$ of the roving ABX task to fit a psychometric function for Proportion Correct (PC)
as in~\cite{freeman2011metamers,Wallis378521}, where the detectability is computed via $d^2(s) = \beta_0(1-\frac{s_o^2}{s^2})\mathbbm{1}_{s> s_0}$, and
\begin{equation}
\label{eq:roving_ABX}
 PC(s) = \Phi\left(\frac{d^2(s)}{\sqrt(6)}\right)\Phi\left(\frac{d^2(s)}{2}\right)+\Phi\left(\frac{-d^2(s)}{\sqrt(6)}\right)\Phi\left(\frac{-d^2(s)}{2}\right)
\end{equation}

\begin{figure}[t]
\centering
\includegraphics[width=1.0\columnwidth,clip=true,draft=false,]{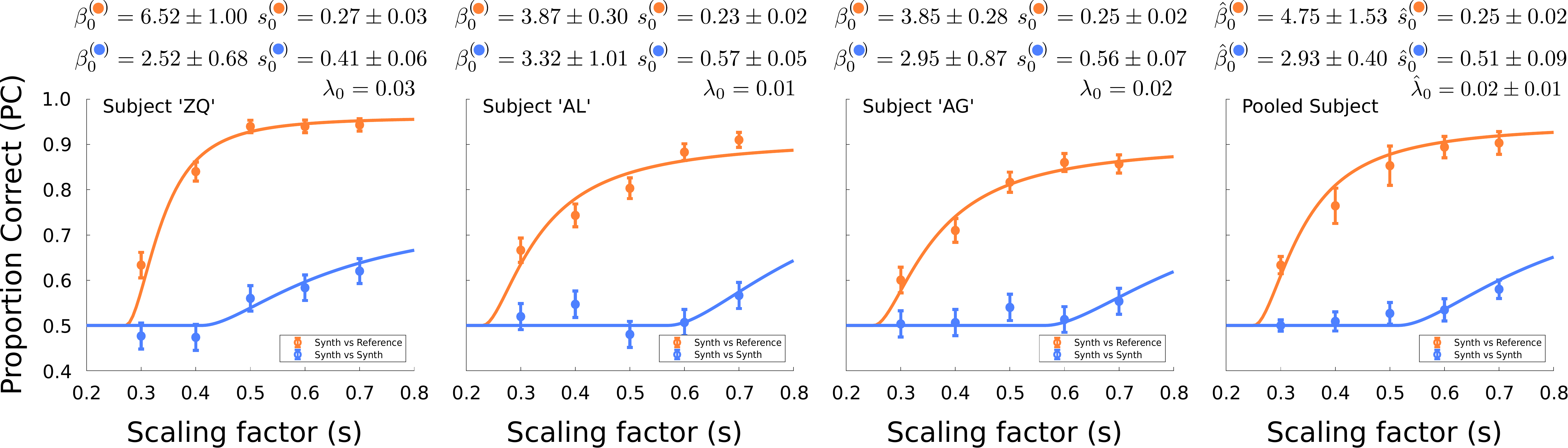}
\vspace{-10pt}
  \caption{The results of the 3 observers and the pooled observer (average; shown on far right) 
  for the Synth vs Reference and Synth vs Synth experiment for our metamers. The error bars
  denote the $68\%$ confidence interval after bootstrapping the trials per observer.
  }\label{fig:Exp2_Results}
\vspace{-10pt}
\end{figure}

\textbf{Results:} 
  Absorbing gain factors $\beta_0$ and critical scales $s_0$ per observer are shown in Figure~\ref{fig:Exp2_Results}, where 
the fits were made using a least squares curve fitting model and bootstrap sampling $n=10000$ times to produce 
the $68\%$ confidence intervals. Lapse rates $(\lambda)$ were also included for robustness of fit as in~\cite{wichmann2001psychometric}.  
Analogous to~\cite{freeman2011metamers},
  we find that the critical scaling factor is 0.51 when doing the Synth vs Synth experiment which match V2, a critical 
  region in the brain that has been identified to respond to texture as in~\cite{Long201719616,ziemba2016selectivity}. 
  This suggests that the parameters we use to capture and transfer texture statistics which are different from the 
  correlations of a steerable pyramid decomposition as proposed in~\cite{portilla2000parametric}, might the match perceptual 
  discrimination rates of the FS metamers. This does not imply that the models are perceptually equivalent, but it aligns with the results
  of~\cite{ustyuzhaninov2017texture} which shows 
  that even a basis of random filters can also capture texture statistics, thus different flavors of metamer models
  can be created with different statistics.
  In addition, we find that the critical scaling factor for the Synth vs Reference experiment
  is less than 0.5 ($\sim0.25$, matching V1) for the pooled observer as validated 
  recently by~\cite{Wallis378521} for their CNN synthesis and FS model for the Synth vs Reference condition. 

\vspace{-12pt}
\section{Discussion}
\vspace{-10pt}

There has been a recent surge in interest with regards to developing and testing new
metamer models: The SideEye model developed 
by ~\cite{fridman2017sideeye}, uses a fully convolutional network (FCN) as in~\cite{long2015fully} and
learns to map an input image into a Texture Tiling Model (TTM) mongrel~(\cite{rosenholtz2012summary}). 
Their end-to-end 
model is also feedforward like ours, but no use of noise is incorporated in the generation pipeline making their model fully
deterministic. At first glance this seems to be an advantage rather a limitation, however it limits the biological plausilibility
of metameric response as the same input image should be able to create more than one metamer. 
Another model
which has recently been proposed is the CNN synthesis model developed by~\cite{Wallis378521}. 
The CNN synthesis model is gradient-descent based and is 
closest in flavor to the FS model, with the difference that their texture statistics are provided by 
a gramian matrix of filter activations of multiple layers of a VGGNet, rather than those used in~\cite{portilla2000parametric}.


The question of whether the scaling parameter is the only parameter to be optimized for metamerism still seems to be open.
This has been questioned early in~\cite{rosenholtz2012summary}, and recently proposed
and studied by~\cite{Wallis378521}, who suggest that metamers are driven by image content, rather than bouma's law
(scaling factor). Figure~\ref{fig:Alpha_Figure} suggests that
on average, it does seem that $\alpha$ must increase 
in proportion to retinal eccentricity, but this is conditioned by the image content of each receptive 
field. We believe that the hyperparametric nature of our model sheds some light into reconciling these two theories. 
Recall that in Figures (\ref{fig:Perceptual_Space},~\ref{fig:Metamers_Optimization}), we found that certain images can be pushed stronger in the direction
of it's texturized version versus others given their location in the encoded space, the local geometry of the surface, and their projection in the perceptual space.
This suggests that the average maximal distortion one can do is fixed 
contingent on the size of the receptive field, but we are allowed to \textit{push further} (increase $\alpha$) for some images more 
than others, because the direction of the distortion lies closer to the perceptual null space (making this difference perceptually un-noticeable
to the human observer). 
This is usually the case
for regions of images that are periodic like skies, or grass.
Along the same lines, we elaborate in the Supplementary Material on how our model
may potentially explain why creating synthesized samples are metameric to each other at the 
scales of (V1;V2), but only generated samples at the scale of V1 $(s=0.25)$ are metameric to
the reference image.

\begin{wrapfigure}{l}{0.40\textwidth}
\centering
\includegraphics[width=0.40\columnwidth,clip=true,draft=false,]{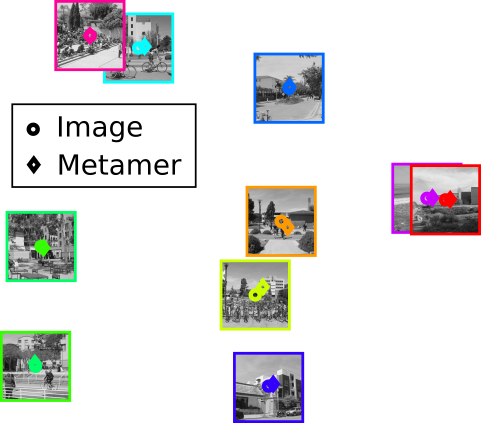}
\vspace{-10pt}
  \caption{Image embeddings.
  }\label{fig:Embedding}
\vspace{-15pt}
\end{wrapfigure}

Our model is also different to others (FS and recently~\cite{Wallis378521}) 
given the role of noise in the computational pipeline.
The previously mentioned models used noise as an initial seed for the texture matching pipeline via gradient-descent,
while we use noise as a proxy for texture distortion that is directly associated with crowding in the visual field.
One could argue that the same response is achieved via both approaches, but our approach seems to be more biologically plausible
at the algorithmic level. In our model an image is fed through a non-linear 
hierarchical system (simulated through a deep-net), 
and is corrupted by noise that matches the texture properties of the input 
image (via AdaIN). This perceptual representation is perturbed along the direction
of the texture-matched patch for each receptive field, and inverting such perturbed representation results in a metamer. 
Figure~\ref{fig:Embedding}
illustrates such perturbations which produce metamers when projected to a 2D subspace via the locally linear 
embedding (LLE) algorithm~(\cite{roweis2000nonlinear}).
Indeed, the 10 encoded images 
do not fully overlap to each other and they are quite distant as seen in the 2D projection.
However, foveated representations when perturbed with texture-like noise 
seem to finely tile the perceptual space, 
and might act as a type of \textit{biological regularizer} for human observers who 
are consistently making eye-movements when processing visual information. This suggests that robust representations
might be achieved in the human visual system given its foveated nature as non-uniform high-resolution imagery does not map
to the same point in perceptual space. 
If this holds, perceptually invariant data-augmentation
schemes driven by metamerism may be a useful enhancement for artificial systems that react oddly to adversarial 
perturbations that exploit coarse perceptual mappings
~(\cite{goodfellow2014explaining,tabacof2016exploring,berardino2017eigen}).


Understanding the underlying representations of metamerism in the human visual system still remains a challenge.
In this paper we propose a model that emulates metameric responses via a foveated feed-forward style transfer network. We find that correctly calibrating such perturbations 
(a consequence of internal noise
that match texture representation) in the perceptual space
and inverting such encoded representation results in a metamer.
Though our model is hyper-parametric in nature we propose a way to reduce the parametrization via a 
perceptual optimization scheme. 
Via a psychophysical experiment we empirically find that the critical scaling factor also matches the rate
of growth of the receptive fields in V2 ($s=0.5$) as in~\cite{freeman2011metamers} 
when 
performing visual discrimination between synthesized metamers, and match V1 (0.25) for reference metamers similar 
to~\cite{Wallis378521}. 
Finally, while our choice of texture 
statistics and transfer is $relu4\_1$ of a VGG19 and AdaIN respectively, our $\times1000$-fold accelerated feed-forward metamer generation 
pipeline should be extendible to other models that correctly compute texture/style statistics and transfer. This opens the door
to rapidly generating multiple flavors of visual metamers with applications in neuroscience and computer vision.



\section*{Acknowledgements}
We would like to thank Xun Huang for sharing his code and valuable suggestions on AdaIN, Jeremy Freeman for making
his metamer code available, Jamie Burkes for collecting original high-quality stimuli, 
N.C. Puneeth for insightful conversations on texture and masking, 
 Christian Bueno for informal lectures on homotopies, and 
 Soorya Gopalakrishnan and Ekta Prashnani for insightful discussions.
Lauren Welbourne, Mordechai Juni, Miguel Lago, and Craig Abbey were also helpful in editing the 
 manuscript and giving positive feedback. 
We would also like to thank NVIDIA for donating a Titan X GPU. 
This work was supported by the Institute for Collaborative Biotechnologies through grant 2 W911NF-09-0001 
from the U.S. Army Research Office.

\newpage

{\footnotesize
\bibliography{iclr2019_conference}
\bibliographystyle{iclr2019_conference}
}

\newpage

\vspace{-10pt}
\section{Supplementary Material}
\vspace{-10pt}

\begin{figure}[h]

\centering
\includegraphics[width=0.95\columnwidth,clip=true,draft=false,]{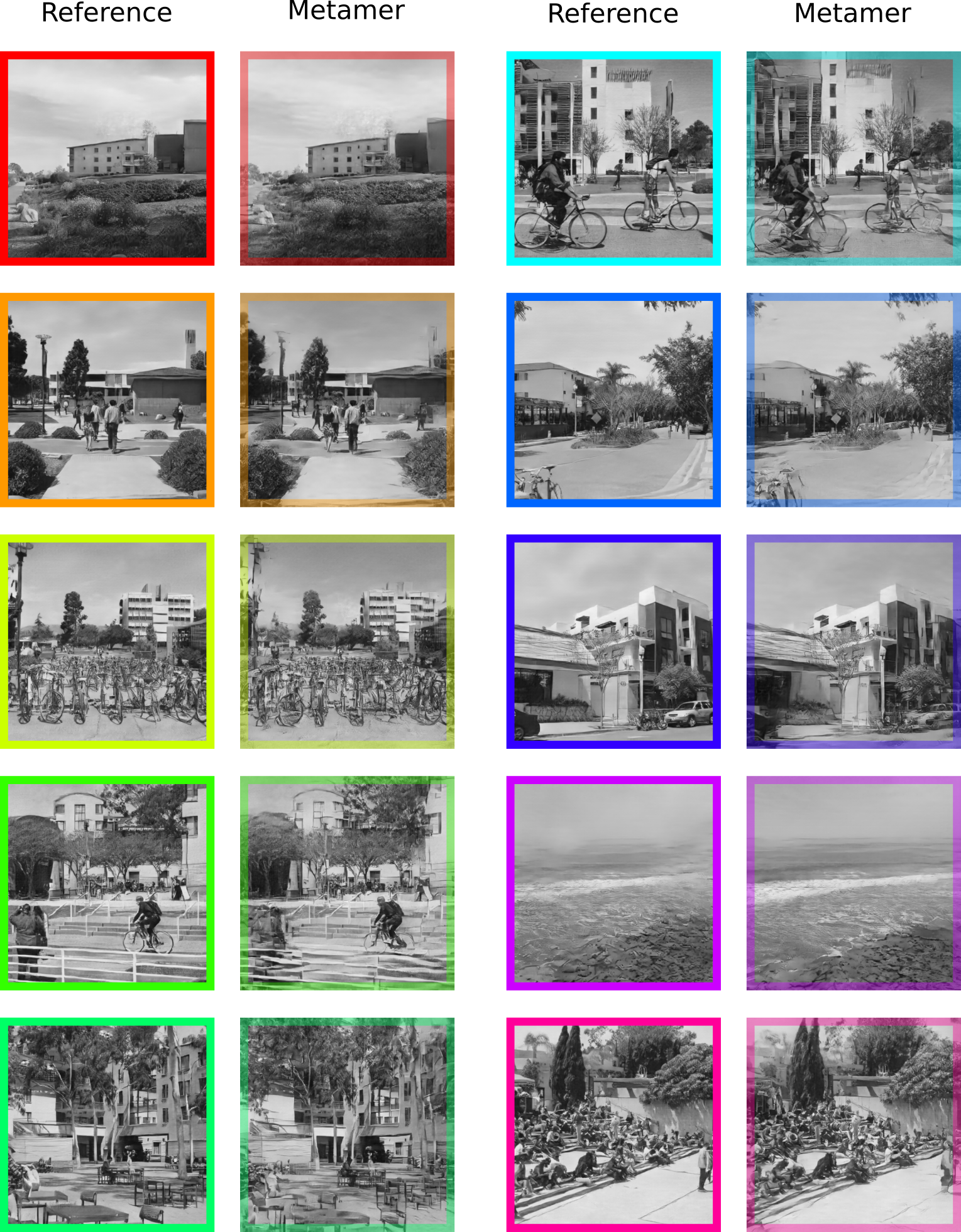}
\vspace{-10pt}
  \caption{Reference Metamers at the scale of $s=0.25$, at which they are indiscriminable to the human observer.
  The color coding scheme matches the data points of the optimization in Experiment 1 and the psychophysics of Experiment 2. 
  All images used in the experiments were generated originally at $512\times512$ px subtending
   $26\times26$  d.v.a (degrees of visual angle).
  }\label{fig:Reference_Metamers_V1}
\vspace{-10pt}
\end{figure}

\newpage

\subsection{Hyperparameter search algorithm}

Algorithm~\ref{alg:Hyperparameter} fully describes the outline of Experiment 1.

\begin{algorithm}[!t]
\small
\caption{Pipeline for Metamer hyperparameter $\gamma(\circ)$ search}
\label{alg:Hyperparameter}
\begin{algorithmic}[1]
\Procedure{Estimate hyperparameter: $\gamma(\circ)$ function}{} \\
Choose image dataset $S_I$.\\
Pick hyperparameter search step size $\alpha_{\text{step}}$.
Pick scale search step size $s_{\text{step}}$.
\For{each image $I \in S_I$}
\For{each scale $s\in [s_{\text{init}}:s_{\text{step}}:s_{\text{final}}]$}
\State Compute baseline metamer $\text{M}_{\text{FS}}(I)$
\For{each $\alpha\in[0:\alpha_{\text{step}}:1]$}
\State Compute metamer $\text{M}_{\text{NF}}(I)$
\EndFor
\State Find the $\alpha$ for each receptive field that minimizes: $\mathbb{E}(\Delta\text{-SSIM})^2$.
\State Fit the $\gamma_s(\circ)$ function to collection of $\alpha$ values.
\EndFor
\EndFor
\\Perform Permutation test on $\gamma_s$ for all $s$.
\If{$\gamma_s$ is independent of $s$}
\State $\gamma_s = \gamma$
\Else{}
\State Perform regression of parameters of $\gamma_s$ as a function $f$ of $s$.
\State $\gamma_s = \gamma_{f(s)}$
\EndIf
\EndProcedure
\end{algorithmic}
\end{algorithm}

\subsection{Model specificiations and training}
We use $k=k_p+k_f$ spatial control windows, $k_p$ pooling regions ($\theta_r$ receptive fields $\times$ $\theta_t$ eccentricity rings)
and $k_f=1$ fovea (at an 
approximate $3\deg$ radius). 
Computing the metamers for the scales of $\{0.3,0.4,0.5,0.6,0.7\}$ required $\{300,186,125,102,90\}$ 
pooling regions excluding the fovea where we applied local style transfer.
Details regarding the decoder network architecture and training can be seen in~\cite{huang2017adain}. 
We used the publicly available code by Huang and Belongie for our decoder which was trained on ImageNet and a collection of publicly available
paintings to learn how to invert texture as well. In their training 
pipeline, the encoder is fixed and the decoder is trained to learn how to invert the structure of the content image,
and the texture of the style image, thus when the content and style image are the same, then the decoder approximates
the inverse of the encoder $(\mathcal{D}\sim\mathcal{E}^{-1})$. 
We also re-trained another decoder on a set of 100 images all being scenes (as a control to check for potential differences), 
and achieved similar outputs (visual inspection) to the publicly available one of Huang~\&~Belongie.
The dimensionality of the 
input of the encoder is $1\times512\times512$, and the dimensionality of the output $(relu4\_1)$ is $512\times64\times64$, it is at the $64\times64$
resolution that we are applying foveated pooling from the initial guidance channels of the $512\times512$ input.

Constructions of biologically-tuned 
peripheral representations are explained in detail in~\cite{freeman2011metamers,akbas2017object,deza2016can}, 
and are governed by the following equations:

\begin{equation}
\label{mod:f}
f(x) = \begin{cases}
        cos^2(\frac{\pi}{2}(\frac{x-(t_0-1)/2}{t_0})); & -(1+t_0)/2<x\leq (t_0-1)/2 \\
        1                                        ; & (t_0-1)/2 < x \leq (1-t_0)/2 \\
        -cos^2(\frac{\pi}{2}(\frac{x-(1+t_0)/2}{t_0}))+1; & (1-t_0)/2 < x \leq (1+t_0)/2
       \end{cases}
\end{equation}

\begin{equation}
\label{mod:h}
 h_n(\theta) = f\Big(\frac{\theta-(w_{\theta}n+\frac{w_{\theta}(1-t_0)}{2})}{w_{\theta}}\Big); w_{\theta} = \frac{2\pi}{N_\theta}; n=0,...,N_\theta-1
\end{equation}

\begin{equation}
\label{mod:g}
 g_n(e) = f\Big(\frac{\log(e)-[\log(e_0)+w_e(n+1)]}{w_e}\Big); w_e = \frac{\log(e_r)-\log(e_0)}{N_e}; n=0,...,N_e-1 
\end{equation}
where $f(x)$ is a cosine profiling function that smoothes a regular step function,
and $h_n(\theta)$,$g_n(e)$, are the averaging values of the pooling region $w_i$ at a specific angle $\theta$ and radial eccentricity
$e$ in the visual field. In addition we used the default values of visual radius of $e_r = 26\deg$, and $e_0 = 0.25\deg$~\footnote{We remove
central regions with an area smaller than 100 pixels, and group them into the fovea}, 
and $t_0 = 1/2$. 
The scale $s$ defines the number of eccentricities $N_e$, as well as the number of polar pooling regions $N_{\theta}$ from
$\langle 0, 2\pi]$. We perform the foveated
pooling operation on the output of the Encoder. Since the encoder is fully convolutional with no fully connected layers,
guidance channels can be used to do localized (foveated) style transfer.

Our pix2pix U-Net refinement module took 3 days to train on a Titan X GPU, and was trained with 64 crops $(256\times256)$ per image on 100 images, including horizontally 
mirrored versions.
We ran 200 training epochs of these 12800 images on the U-Net architecture proposed 
by~\cite{isola2016image} which preserves 
local image structure given an adversarial and L2 loss.  

\subsection{Metamer Model Comparison}
The following table summarizes the main similarities and differences across all current models:

\begin{table}[h]
\scriptsize
\centering
\begin{tabular}{|c|c|c|c|c|}
 \hline
Model & FS (2011) & CNN-Synthesis (2018) & SideEye (2017) & NF (Ours) \\
 \hline
Feed-Forward       & - & - & \checkmark & \checkmark  \\
Input & Noise & Noise & Image & Image \\
Multi-Resolution & \checkmark & \checkmark & - & - \\
Texture Statistics & Steerable Pyramid & VGG19 \textit{conv-}$1_1,2_1,3_1,4_1,5_1$ & Steerable Pyramid & VGG19 \textit{relu}$4_1$ \\
Style Transfer & Portilla~\&~Simoncelli & Gatys et al. & Rosenholtz et al. & Huang~\&~Belongie\\
Foveated Pooling & \checkmark & \checkmark & (Implicit via FCN) & \checkmark \\
Decoder (trained on) & - & - & metamers/mongrels & images \\ 
Moveable Fovea & \checkmark & \checkmark & \checkmark & \checkmark \\
\hline
Use of Noise & Initialization & Initialization & -  & Perturbation \\
Non-Deterministic & \checkmark & \checkmark & -  & \checkmark \\
Direct Computable Inverse & - & - & (Implicit via FCN) & \checkmark \\
\hline
Rendering Time & hours & minutes & miliseconds & seconds \\
\hline
Image type & scenes & scenes/texture & scenes & scenes\\ 
Critical Scaling (\textit{vs} Synth) & 0.46 & $\sim\{0.39/0.41\}$ & Not Required & 0.5 \\
Critical Scaling (\textit{vs} Reference) & Not Available & $\sim\{0.2/0.35\}$ & Not Required & 0.24 \\
Experimental design & ABX & Oddball & - & ABX \\ 
Reference Image in Exp. & Metamer & Original & - & Compressed via Decoder \\
Number of Images tested & 4 & 400 & - & 10 \\
Trials per observers & $\sim 1000$ & $\sim 1000$ & - & $\sim 3000$ \\
\hline
\end{tabular}
\caption{Metamer Model comparison}
\label{table:Metamer_Comparison}
\end{table}

\begin{figure}[h]
\centering
\includegraphics[width=1.0\columnwidth,clip=true,draft=false,]{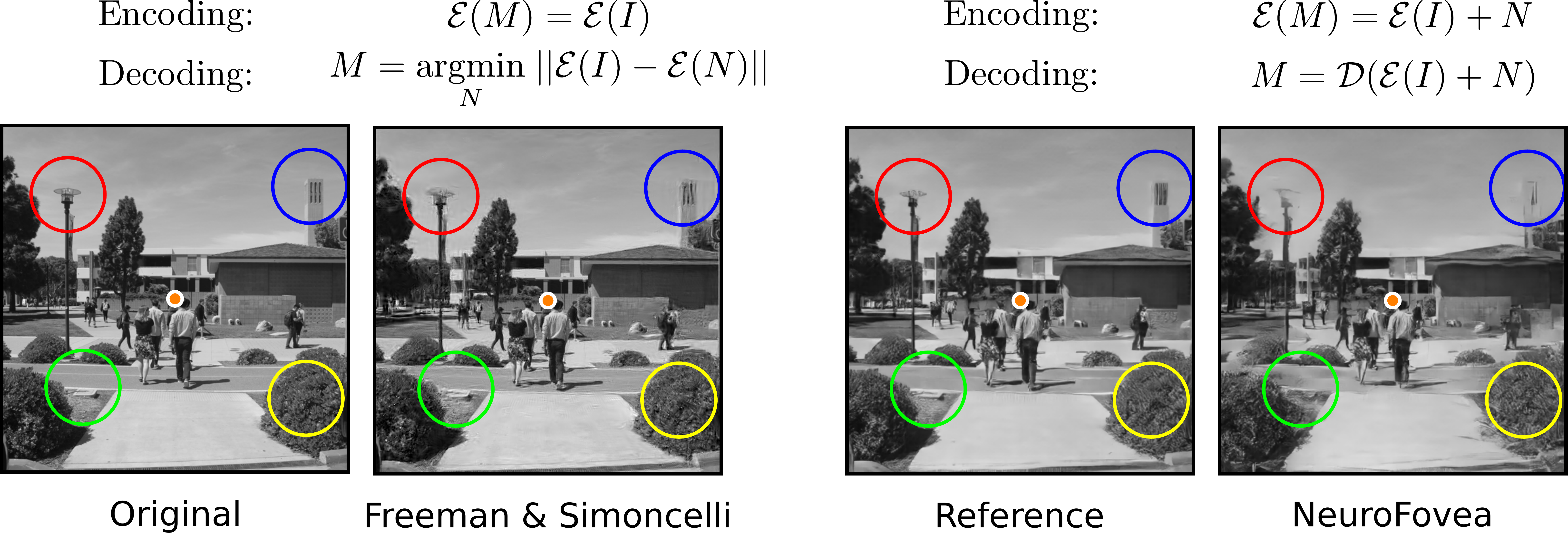}
\vspace{-10pt}
  \caption{Algorithmic (top) and visual (bottom) comparisons between our metamers and a sample from~\cite{freeman2011metamers} 
  for a scaling factor of 0.3. 
  Each model has it's own limitations: The FS model can not directly compute an inverse of the encoded representation to generate
  a metamer, requiring an iterative gradient descent procedure. 
  Our NF model is limited by the capacity of the encoder-decoder architecture as it does not achieve lossless compression (perfect reconstruction).
  }\label{fig:Method_Comparison}
\vspace{-10pt}
\end{figure}

\newpage

\subsection{Interpretability of V1 and V2 Metamers}

\begin{figure}[!t]
\centering
\includegraphics[width=1.0\columnwidth,clip=false,draft=false,]{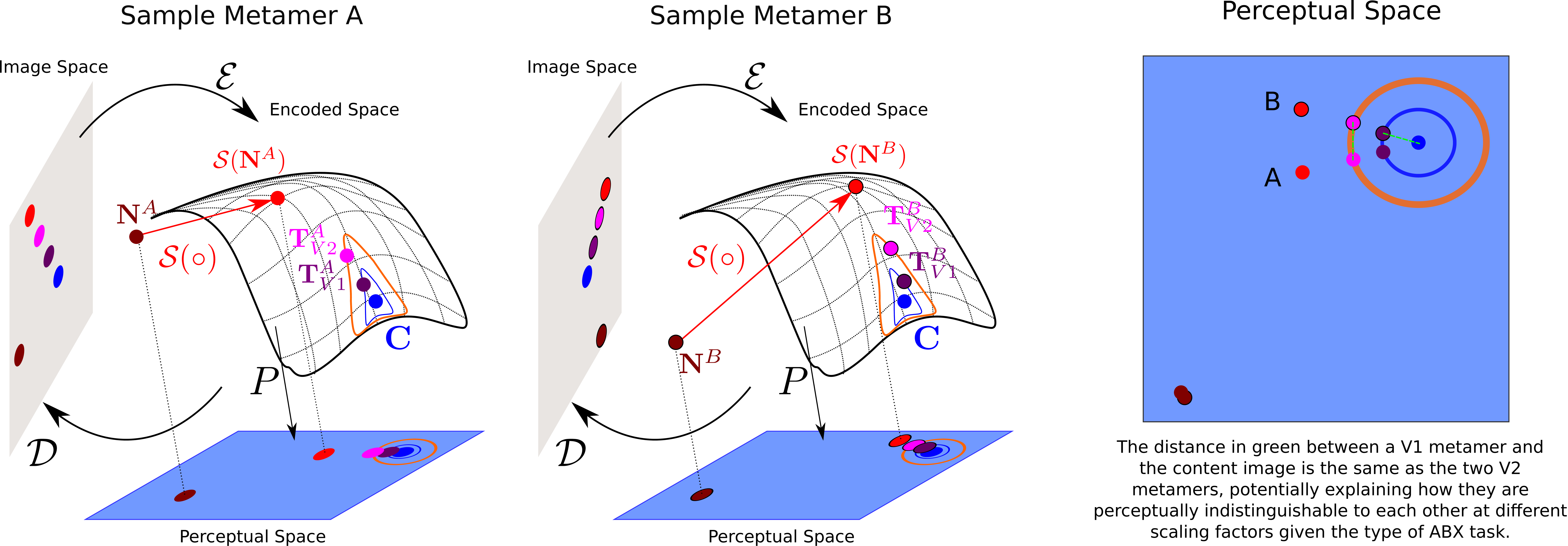}
\vspace{-10pt}
  \caption{Decomposition and overview of the metamer generation process in the Image space, 
  the Encoded space and the Perceptual space. The original image patch is coded in blue, the V1 metamers are coded in purple, and the V2 metamers
  are coded in pink. Dark brown represents the initial white noise that is later stylized via AdaIN through $\mathcal{S}(\circ)$.
  Note that these two points are far away to each other in image space, but quite closeby in perceptual space as they are also 
  `metameric' to each other. They are not placed on the actual encoded manifold since these points are not in the near vicinity 
  of either
  $\mathbf{C}$ nor $\mathcal{S}(\mathbf{N})$, as they have no scene-like structure.
  The interpolation for maximal distortion 
  is done along the line between $\mathbf{C}$ and $\mathcal{S}(\mathbf{N})$, these are the points in blue and red in the encoded space which represent
  the extremes of $\alpha=0.0$ and $\alpha=1.0$ respectively.
  }\label{fig:Perceptual_Manifolds_Detail}
\vspace{-13pt}
\end{figure}

In Figure~\ref{fig:Perceptual_Manifolds_Detail}, we illustrate 
the metamer generation process for two sample metamers, given different
noise perturbations. Here, we
decompose Figure~\ref{fig:Perceptual_Space} into two separate ones for each metamer given each noise perturbation, 
and provide an additional visualization of the projection of the metamers in
perceptual space, gaining theoretical insight on how and why metamerism arises for the synth-vs-synth condition in V2,
and the synth-vs-reference condition in V1 as we demonstrated experimentally.

\subsection{Pilot Experiments}
\begin{wrapfigure}{l}{0.45\textwidth}
\centering
\includegraphics[width=0.40\columnwidth,clip=true,draft=false,]{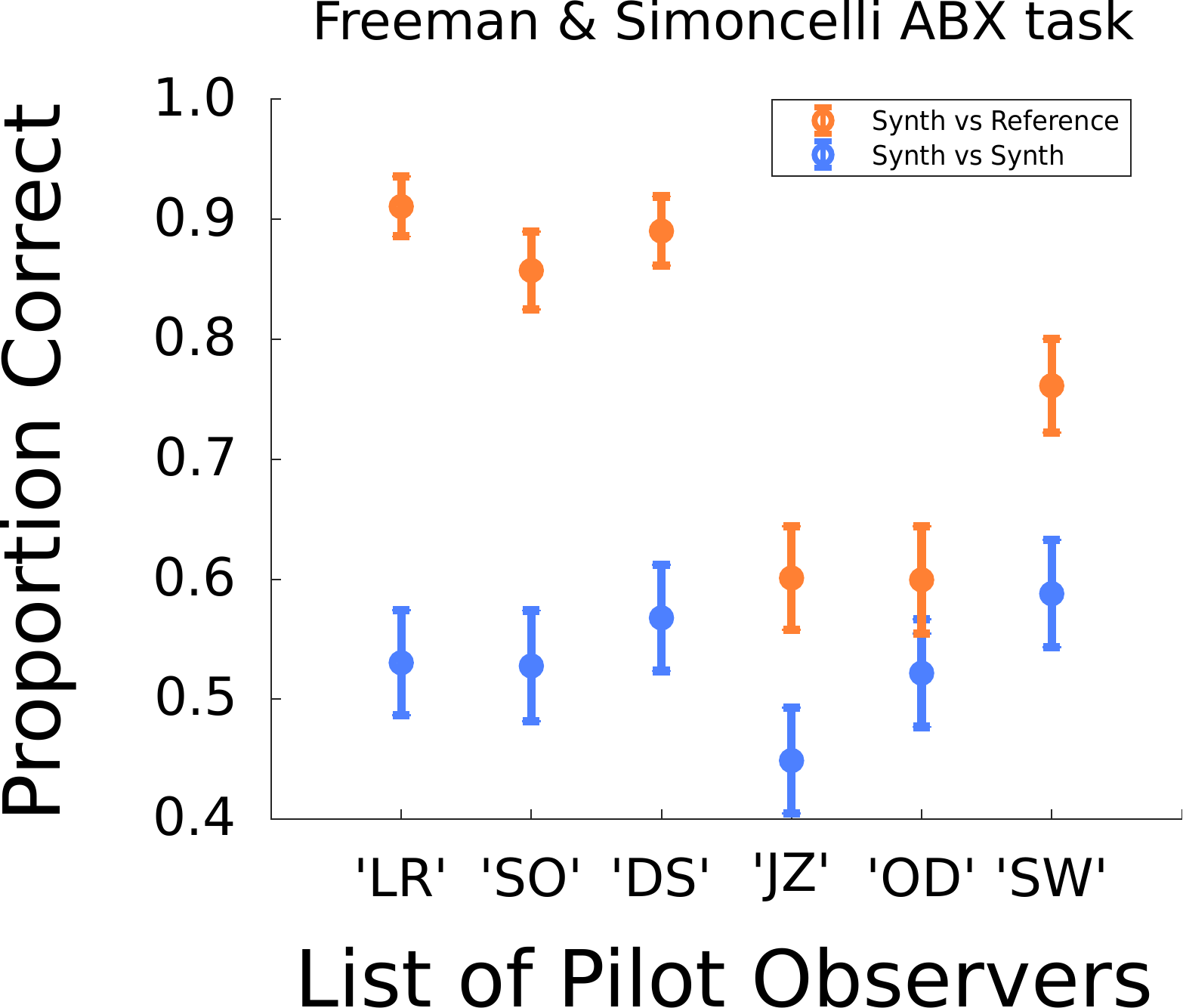}
\vspace{-10pt}
  \caption{Pilot Data on FS metamers.
  }\label{fig:Pilot_Data}
\vspace{-10pt}
\end{wrapfigure}
In a preliminary psychophysical study, we ran an experiment with a collection of 50 images and 6 observers on the FS 
metamers. Observers performed a single session of 200 trials of the FS metamers where the scale was fixed at $s=0.5$.
We found the following: 
While we found that the synthesized images were metameric to each other for the scaling factor of $0.5$,
the FS metamers were not metameric to their reference high-quality images at the scale of $0.5$.
Only a sub-group of observers: \textit{`LR'},\textit{`SO'},\textit{`DS'} scored well above chance in terms of discriminating the images in the ABX task.
These results are in synch with the evalutions
done by~\cite{Wallis378521}, which varied scale and found a critical value to be less than $0.5$ and rather closer
to $0.25$ within the range of V1.

\newpage
\subsection{Estimation of Lapse-rate $(\lambda)$ per observer}

The motivation behind estimating the lapse rate is to quantify how engaged was the observer in the experiment, 
as well as providing a robust estimate of the parameters in the fit of the psychometric functions. Not accounting for lapse rate
may dramatically affect the estimation of these parameters as suggested in~\cite{wichmann2001psychometric}.
In general lapse rates are computed by penalizing a psychometric function $\psi(\circ)$ that ranges between some lower bound and upper bound
usually $[0,1]$. To estimate the lapse rate $\lambda$, a new $\psi'(\circ)$ is defined to have the following form:

\begin{equation}
 \psi'(\circ) = b+(1-b-\lambda)\psi(\circ)
\end{equation}

Recall that for us, our psychometric fitting function $\psi(\circ)=PC_{ABX}(s)$ is defined by Equation~\ref{eq:roving_ABX} and parametrized by both
the absorbing factor $\beta_0$
and the critical scaling factor $s_0$:

\begin{equation}
 PC_{ABX}(s) = \Phi\left(\frac{d^2(s)}{\sqrt(6)}\right)\Phi\left(\frac{d^2(s)}{2}\right)+\Phi\left(\frac{-d^2(s)}{\sqrt(6)}\right)\Phi\left(\frac{-d^2(s)}{2}\right)
\end{equation}

where we have:

\begin{equation}
d^2(s) = \beta_0(1-\frac{s_o^2}{s^2})\mathbbm{1}_{s> s_0}
\end{equation}

To compute the new $\psi'(\circ)$, we notice first that our $\psi$ is bounded between $[0.5,1]$, and that
the new $\psi'$ will be a linear combination of a correct guess for a lapse, and a correct decision for a non-lapse 
from which we obtain:

\begin{equation}
PC(s) = \lambda + (1-2\lambda)PC_{ABX}(s)
\end{equation}

as derived in~\cite{Henaff-phd} which includes lapse rates for an AXB task. 
When fitting the curves for each of the $n=10000$ bootstrapped samples, 
we restricted the lapse rate to vary between
$\lambda=[0.00,0.06]$ as suggested in~\cite{wichmann2001psychometric}, and found the following lapse rates:

Observer 1: $\lambda_{ZQ}^{RS}=0.0248\pm0.0209$, $\lambda_{ZQ}^{SS}=0.0430\pm0.0228$.

Observer 2: $\lambda_{AL}^{RS}=0.0008\pm0.0062$, $\lambda_{AL}^{SS}=0.0166\pm0.0215$.

Observer 3: $\lambda_{AG}^{RS}=0.0141\pm0.0243$, $\lambda_{AG}^{SS}=0.0218\pm0.0236$.

We later averaged these lapse rates as there is an equal probability of each type 
of trial to appear (Synth vs Synth, or Reference vs Synth), and refitted each curve 
with the new pooled lapse rate estimates $\lambda'$. Indeed, 
each observer did both experiments in a roving paradigm, rather than doing one experiment after the other -- 
thus we should only have \textit{one} estimate for lapse rate per observer. It is worth mentioning that re-performing
the fits with separate lapse rates did not significantly affect the estimates of critical scaling values, as one might argue
that higher lapse rates will significantly move the critical scaling factor estimates. This is not the case as 
the absorbing factor $\beta$ does not place an upper bound for the psychometric function at $1$.

Our critical estimates of lapse rates were:
$\lambda_{ZQ}=0.0339$,
$\lambda_{AL}=0.0087$,
$\lambda_{AG}=0.0179$, as shown in Figure~\ref{fig:Exp2_Results}. 

The estimates (critical scale $(s_0)$, absorbing factor $(\beta_0)$ and lapse rate $(\lambda_0)$) shown for the pooled observer were obtained by averaging the estimates
over the 3 observers.
\newpage

%
%
%
%
%
%
%
%
%

\newpage
\subsection{Robustness of estimation of $\gamma$ function}

In this subsection we show how the perceptual optimization pipeline is robust to a selection of IQA metrics such as MS-SSIM 
(multi-scale SSIM~\footnote{scale in the context of SSIM is referred to resolution (as in scales of a laplacian pyramid), and is not to be
confused with the scaling factor $s$ of our experiments which encode the rate of groth of the receptive fields.})
from~\cite{wang2003multiscale} and IW-SSIM (information content weighted SSIM) from~\cite{wang2011information}. 

There are 3 key observations that stem from these additional results:
\begin{enumerate}
\item The sigmoidal natural of the $\gamma$ function is found again and is also scale independent, showing the broad applicability of our perceptual optimization 
scheme and how it is extendable to other IQA metrics that satisfy SSIM-like properties (upper bounded, symmetric and unique maximum).
\item The tuning curves of MS-SSIM and IW-SSIM look almost identical, given that IW-SSIM is not more than a weighted version 
of MS-SSIM where the weighting function is the mutual information between the encoded representations of the reference and distortion image across multiple
resolutions. Differences
are stronger in IW-SSIM when the region over which it is evaluated is quite large (\textit{i.e.} an entire image), 
however given that our pooling regions are quite small in size, the IW-SSIM score
asymptotes to the MS-SSIM score. In addition both scores converge to very similar values given that we are averaging these scores over the images and over all the pooling regions that lie within
the same eccentricity ring. We found that $\sim90\%$ of the maximum $\alpha$'s had the same values given the 20 point sampling grid that we 
use in our optimization. Perhaps a different selection of IW hyperparameters (we used the default set),
finer sampling schemes for the optimal value search, as well as averaging over more images, may produce visible differences between both metrics.
\item The sigmoidal slope is smaller for both IW-SSIM and MS-SSIM \textit{vs} SSIM, which yields more conservative distortions (as $\alpha$ is smaller for each receptive field). 
This implies that the model can still create metamers at the estimated found scaling factors of $0.21$ and $0.50$, however they may have
different \textit{critical} scaling factors for the reference vs synth experiment, and for the synth vs synth experiment. 
Future work should focus on psychophysically 
finding these critical scaling factors, and if they still
are within the range of rate of growth of receptive field sizes of V1 and V2.
\end{enumerate}


\begin{figure}[!h]
\centering
\includegraphics[width=1.0\columnwidth,clip=true,draft=false,]{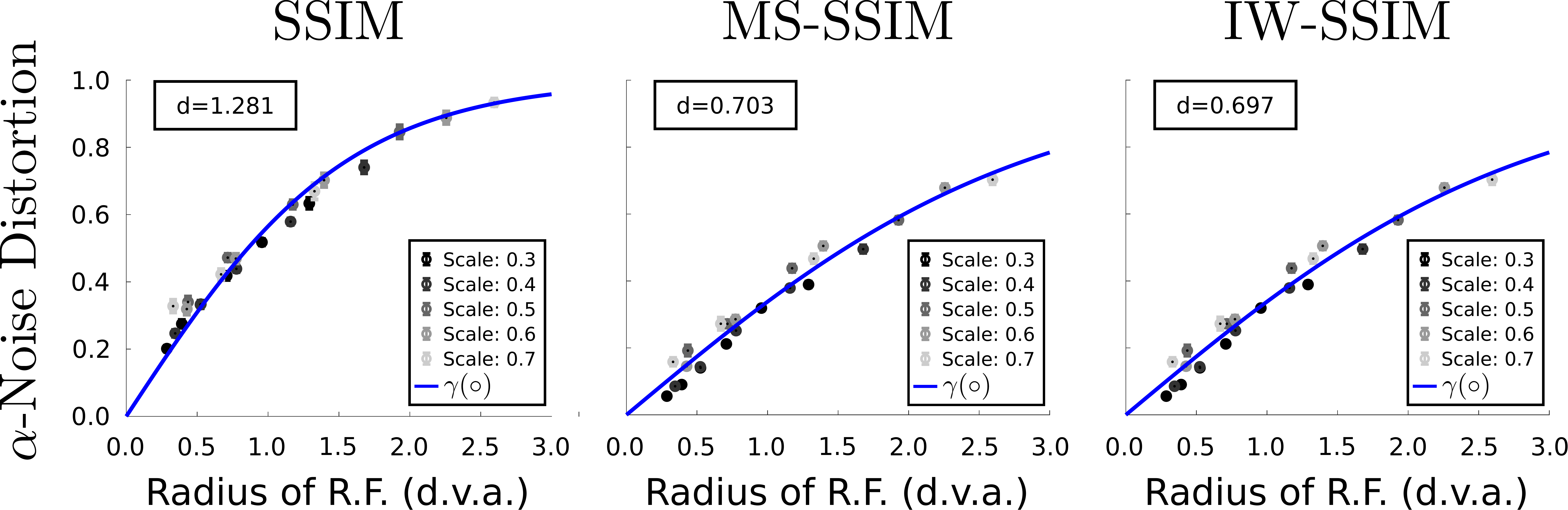}
\caption{A collection of scale invariant $\gamma(\circ)$'s across multiple IQA metrics for the perceptual optimization scheme of Experiment 1.
In this figure we superimpose all maximal $\alpha$-noise distortions for each scale, and find a function that fits all the points showing that
$\gamma$ is indepedent of scale.}
\label{fig:All_Scales_AllSSIM}
\end{figure}

\newpage

\begin{figure}[!h]
\centering
\subfigure[Perceptual Optimization with SSIM.]{
    \includegraphics[width=0.9\columnwidth,clip=true,draft=false,]{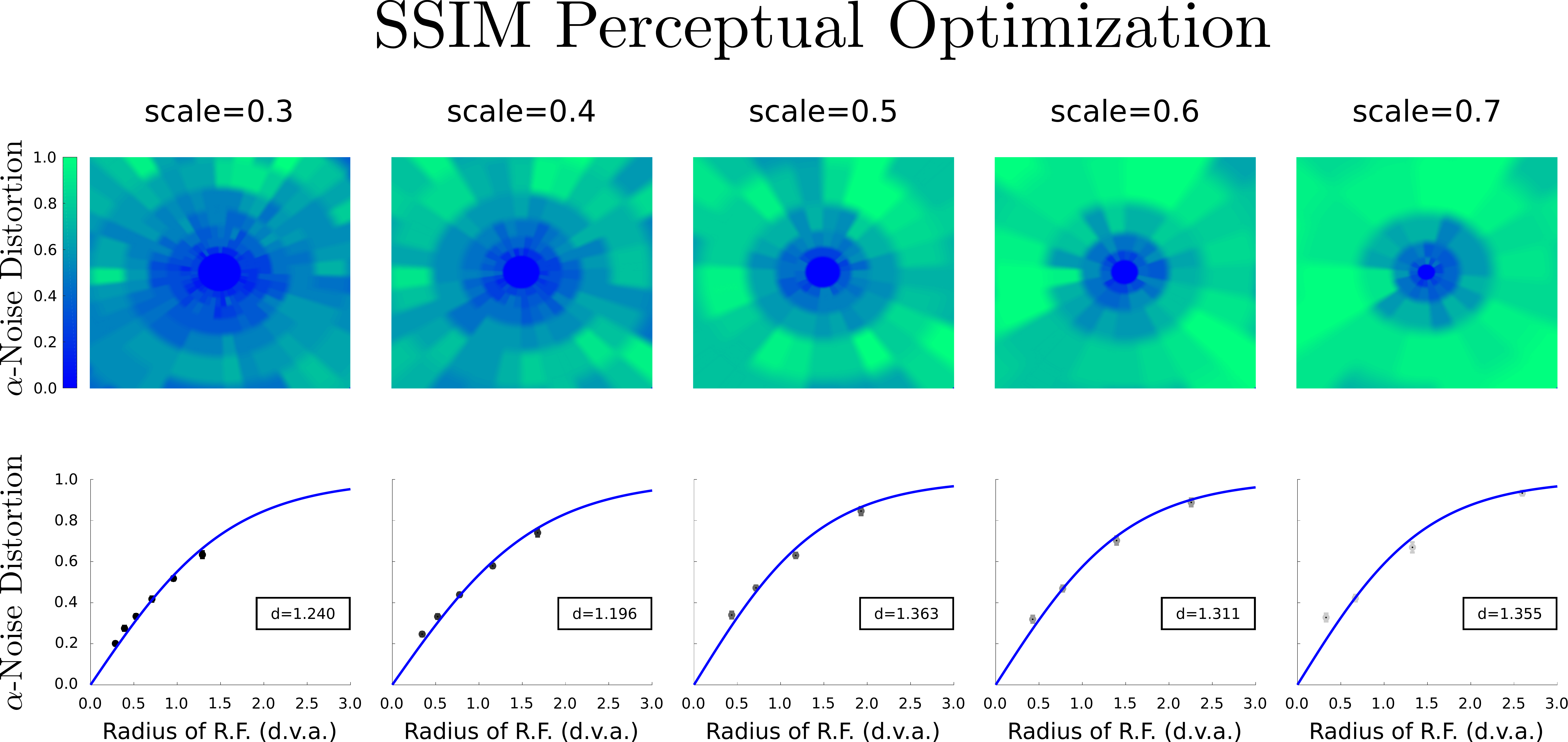}
    \label{fig:Gamma_Func}
}
\subfigure[Perceptual Optimization with MS-SSIM.]{
    \includegraphics[width=0.9\columnwidth,clip=true,draft=false,]{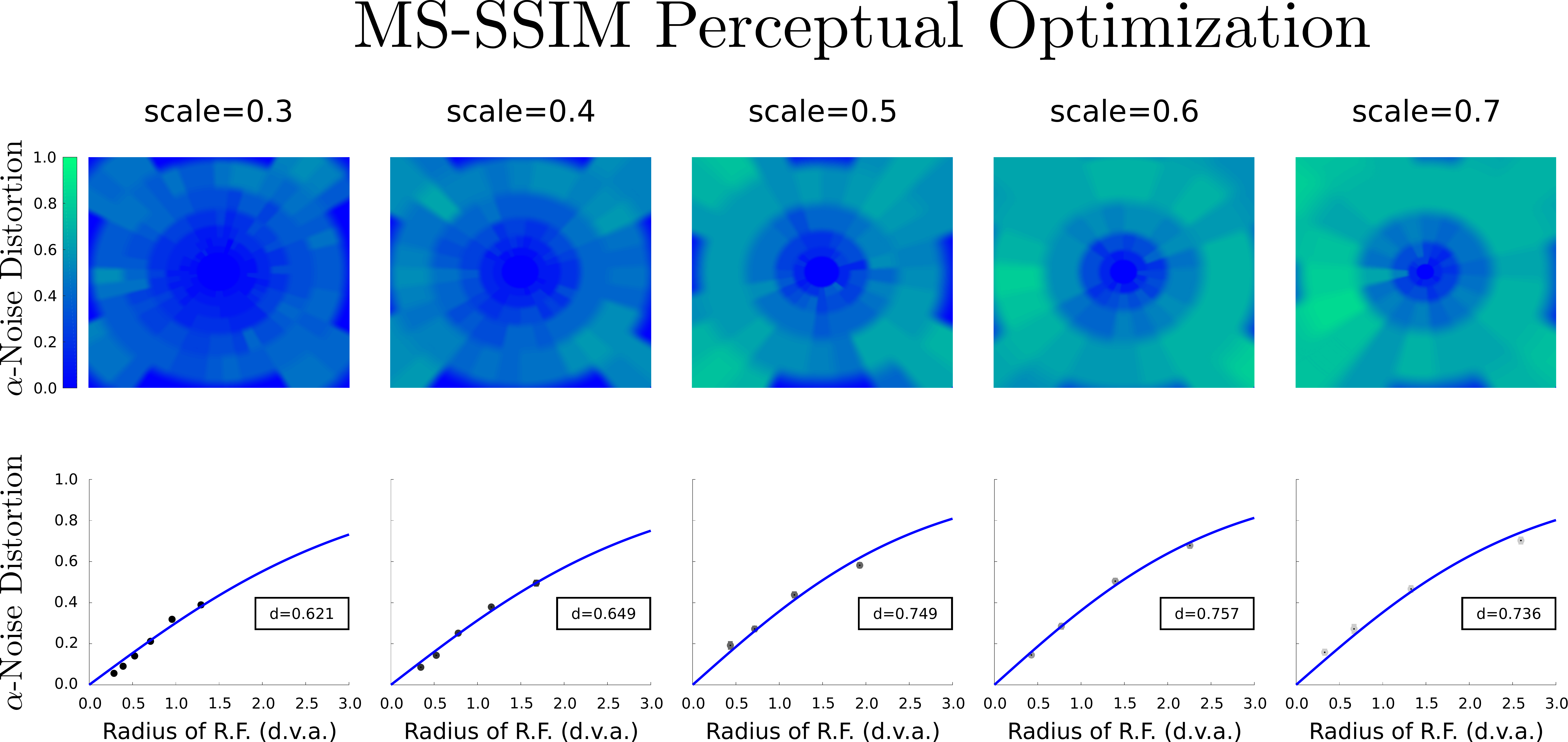}
    \label{fig:Variational}
}
\subfigure[Perceptual Optimization with IW-SSIM.]{
    \includegraphics[width=0.9\columnwidth,clip=true,draft=false,]{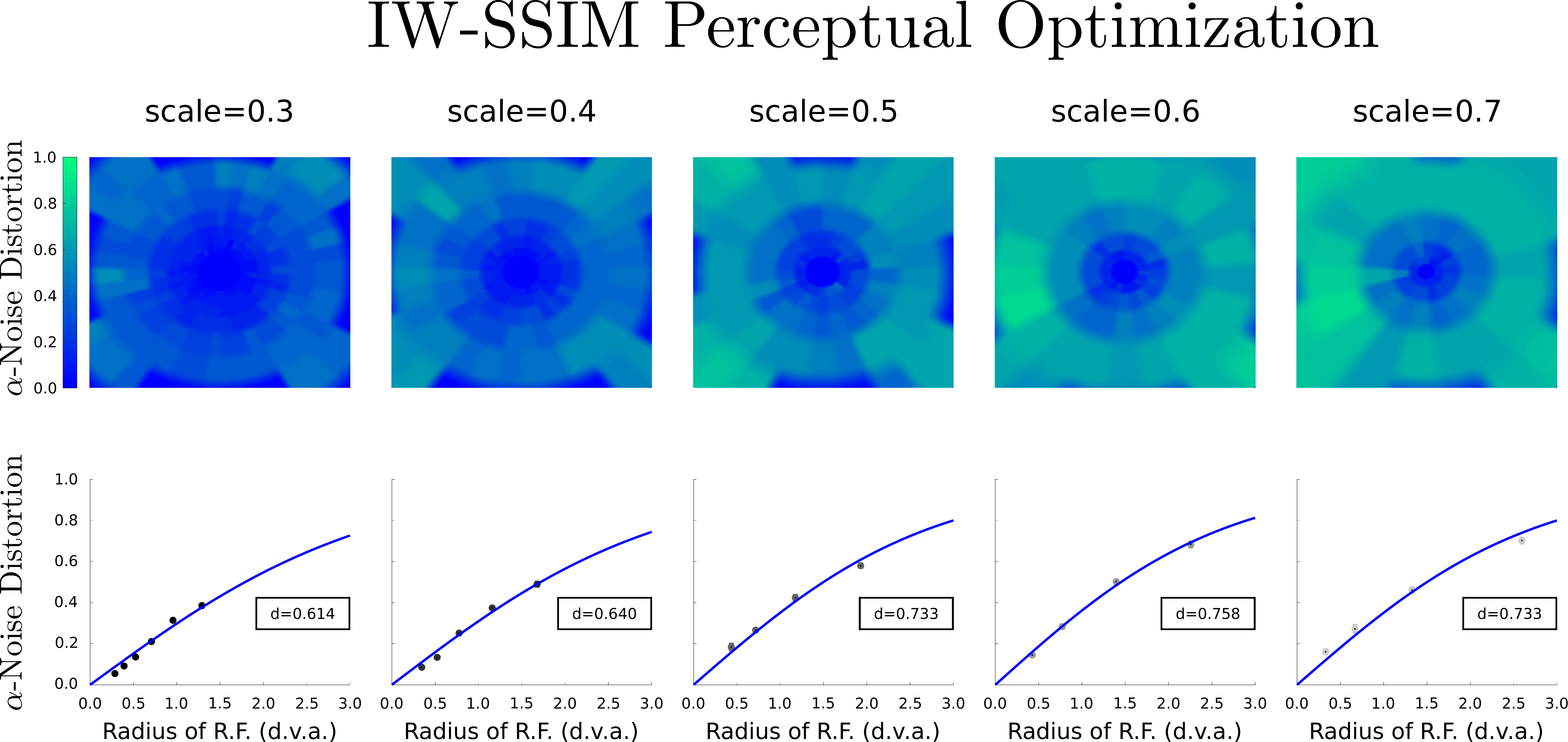}
    \label{fig:Variational}
}
\caption{Top: The maximum $\alpha$-noise distortion computed per pooling region, and collapsed over all images for each IQA metric. 
Bottom: When averaging across all the pooling regions
for each retinal eccentricity, we find that the $\gamma$ function is invariant to scale as in our original experiment -- suggesting that our perceptual optimization
scheme is flexible across IQA metrics.
}\label{fig:Exp1_Results_Extended}
\vspace{-15pt}
\end{figure}


\newpage

\begin{figure}[!h]
\centering
\includegraphics[width=1.0\columnwidth,clip=true,draft=false,]{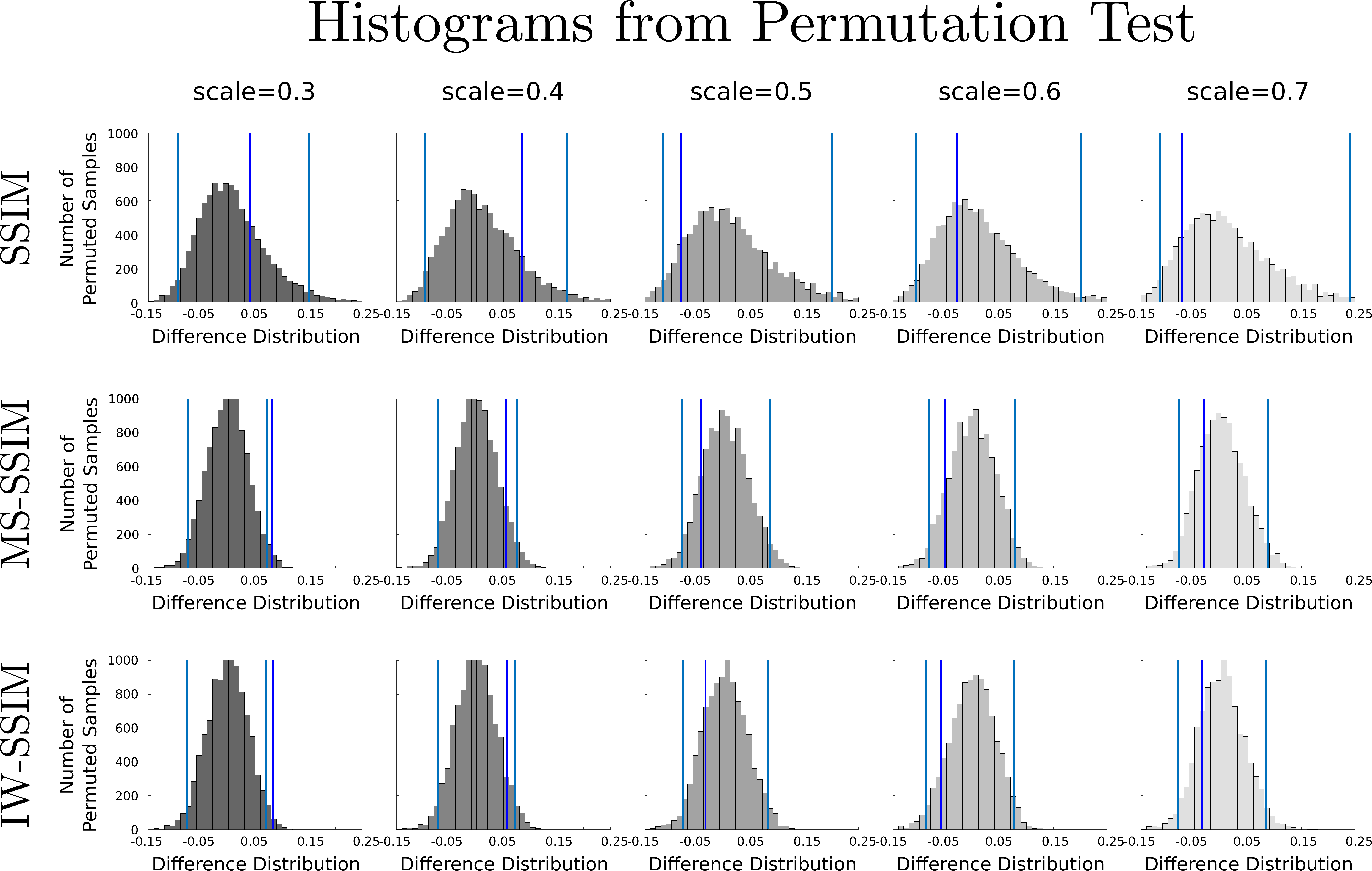}
\caption{A permutation test was ran and determined
that each $\gamma$ function is also scale independent under the $99\%$ confidence interval (CI), as we increased the CI to account for 
false discovery rates (FDR). Indeed, when we perform the permutation tests
and use a $95\%$ confidence interval (shown in the figure with the vertical lines in cyan), all curves except for MS-SSIM and IW-SSIM only for the scaling factor of 0.3
show a significant difference $p\sim0.02$ (non FDR-corrected), potentially due to small receptive field sizes, which bias the estimates. 
All other differences in the $d$ parameter of the sigmoid function, with respect to the average fitted sigmoid, are statistically insignificant.}
\label{fig:All_Permutations}
\end{figure}

\end{document}